\documentclass{article}

\usepackage{arxiv}

\usepackage[utf8]{inputenc} 
\usepackage[T1]{fontenc}    
\usepackage{hyperref}       
\usepackage{url}            
\usepackage{booktabs}       
\usepackage{amsfonts}       
\usepackage{nicefrac}       
\usepackage{microtype}      
\usepackage{lipsum}		
\usepackage{graphicx}
\usepackage{natbib}
\usepackage{doi}
\usepackage{amsmath}
\usepackage{subcaption}
\usepackage{float}  
\usepackage{placeins}   
\usepackage{multirow}
\usepackage{makecell}

\title{DreamHome-Pano: Design-Aware and Conflict-Free Panoramic Interior Generation}

\author{
    Lulu Chen\thanks{Equal contribution.} , 
    Yijiang Hu\footnotemark[1] , 
    Yuanqing Liu, 
    Yulong Li, 
    Yue Yang\thanks{Corresponding author.} \\
    \vspace{2mm} 
    Beike \\
    \texttt{\{chenlulu018, huyijiang001, liuyuanqing006, liyulong008, yangyue092\}@ke.com}
}

\renewcommand{\headeright}{Technical Report}
\renewcommand{\undertitle}{Technical Report}
\renewcommand{\shorttitle}{\textit{DreamHome-Pano}}

\begin{document}
\maketitle

\begin{abstract}
    In modern interior design, the generation of personalized spaces frequently necessitates a delicate balance between rigid architectural structural constraints and specific stylistic preferences. However, existing multi-condition generative frameworks often struggle to harmonize these inputs, leading to "condition conflicts" where stylistic attributes inadvertently compromise the geometric precision of the layout. To address this challenge, we present \textbf{DreamHome-Pano}, a controllable panoramic generation framework designed for high-fidelity interior synthesis. Our approach introduces a \textbf{Prompt-LLM} that serves as a semantic bridge, effectively translating layout constraints and style references into professional descriptive prompts to achieve precise cross-modal alignment. To safeguard architectural integrity during the generative process, we develop a \textbf{Conflict-Free Control} architecture that incorporates structural-aware geometric priors and a multi-condition decoupling strategy, effectively suppressing stylistic interference from eroding the spatial layout. Furthermore, we establish a comprehensive panoramic interior benchmark alongside a multi-stage training pipeline, encompassing progressive Supervised Fine-Tuning (SFT) and Reinforcement Learning (RL). Experimental results demonstrate that DreamHome-Pano achieves a superior balance between aesthetic quality and structural consistency, offering a robust and professional-grade solution for panoramic interior visualization.
\end{abstract}

\begin{figure}[h]
\centering
    \includegraphics[scale=0.4]{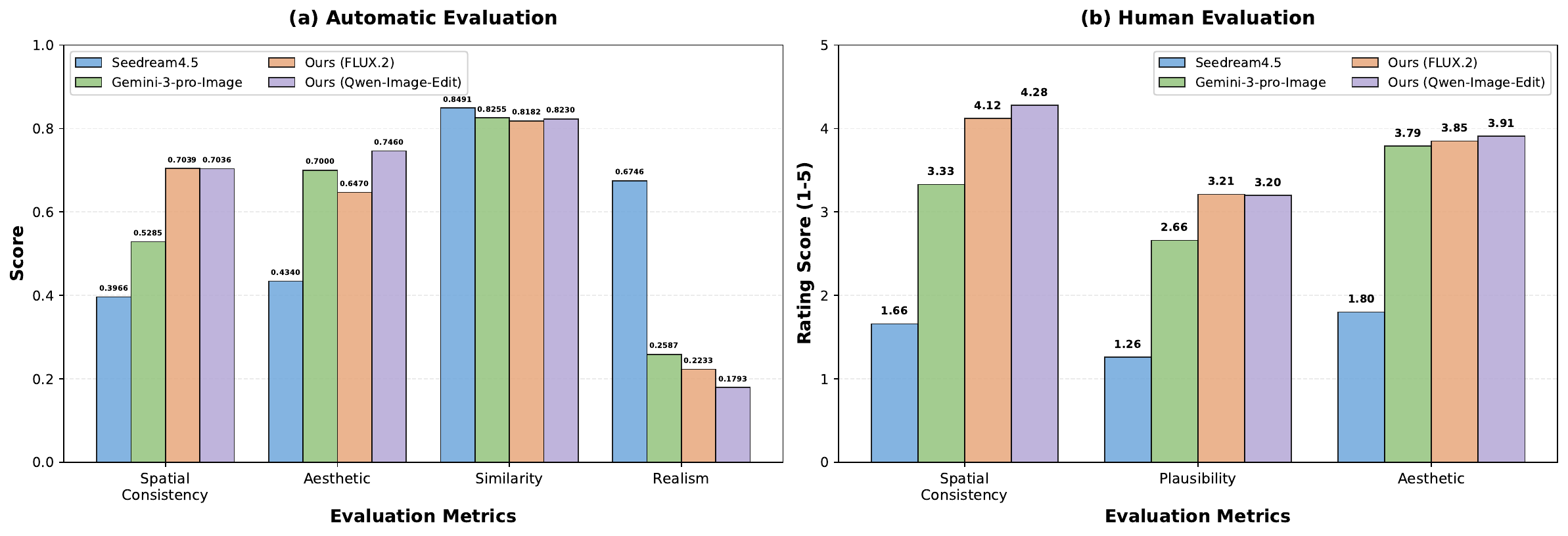}
    \caption{Evaluation of DreamHome-Pano. (a) Automatic evaluation; (b) Human evaluation.}
    \label{fig:result}
\end{figure}

\begin{figure}
\centering
    \includegraphics[scale=0.8]{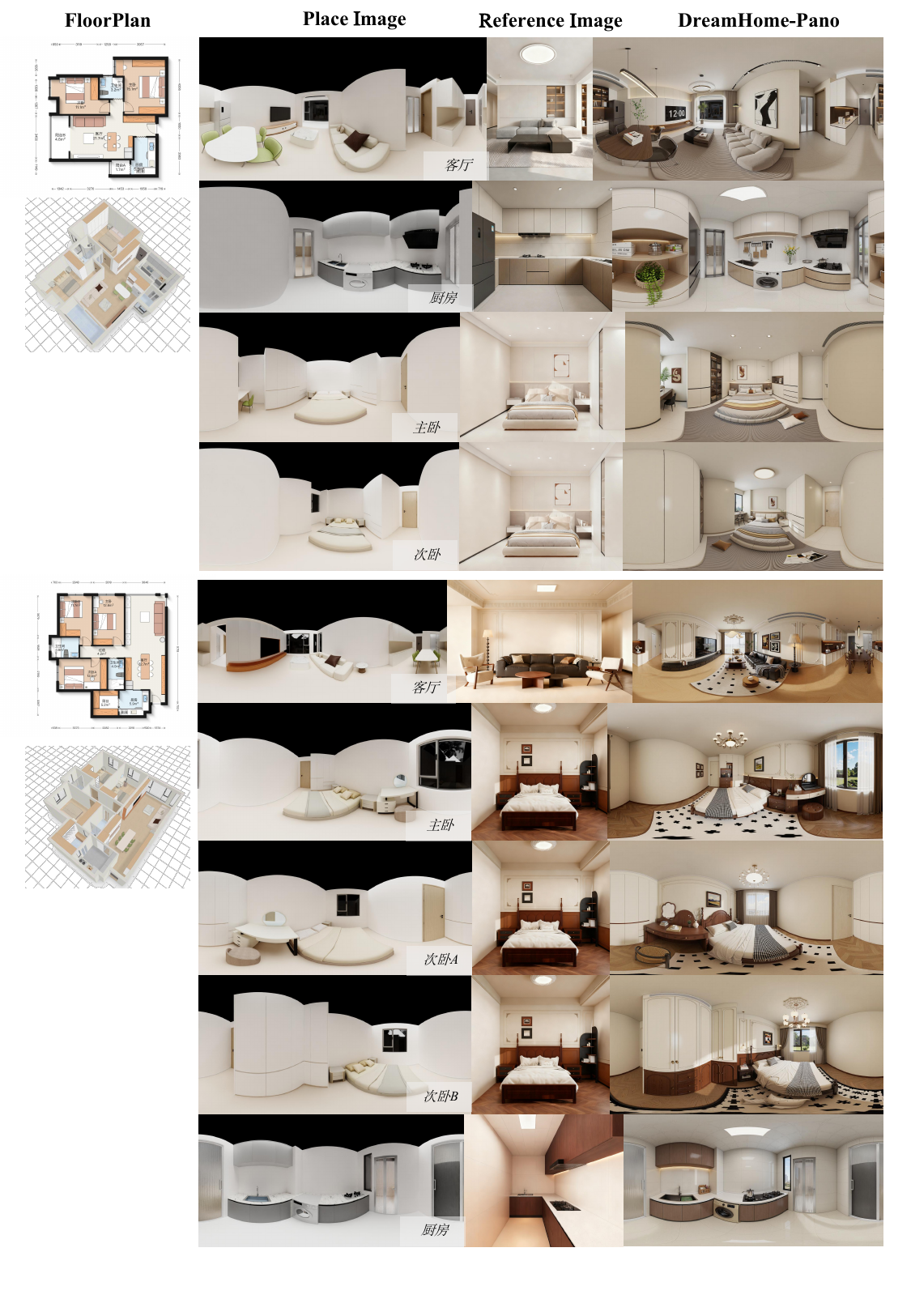}
    \caption{Showcase of panoramic generation guided by floorplans and style reference images.}
    \label{fig:ref-showcase}
\end{figure}

\begin{figure}
\centering
    \includegraphics[width=1\linewidth]{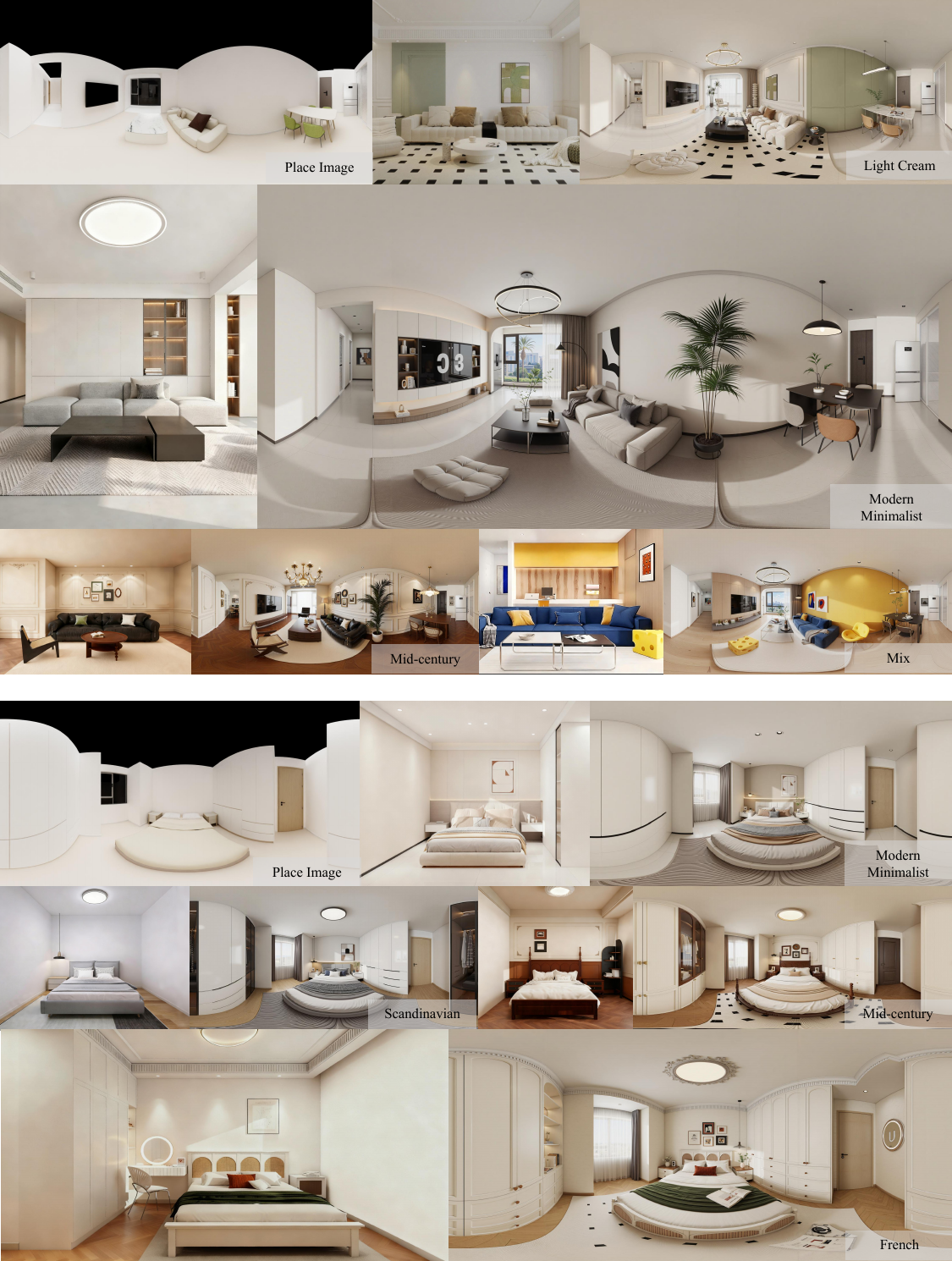}
    \caption{Showcase of the multi-style generation capability of DreamHome-Pano.}
    \label{fig:showcase-result}
\end{figure}

\section{Introduction}
The rapid advancement of generative AI has revolutionized the field of interior design, shifting the paradigm from labor-intensive 3D modeling to efficient generative workflows. In particular, 360-degree panoramic generation has garnered significant attention due to its ability to provide immersive, VR-ready experiences that traditional 2D perspective views cannot offer. 

However, generating truly personalized spaces typically involves two primary visual inputs: a structural layout of the residence and a style reference image representing the desired aesthetic, often accompanied by a textual description. In practical interior design pipelines, the structural layout is not provided as an abstract floorplan, but is instead rendered into a \textbf{Place Image} by placing simplified 3D furniture proxies and basic architectural elements according to the layout and camera configuration. This rendered place image serves as a geometric and spatial placeholder, encoding room boundaries, furniture occupancy, and viewpoint alignment without specifying fine-grained appearance details.

As a result, panoramic interior generation naturally formulates a multi-image editing task, where the model must jointly leverage a layout-based structural control image, a style reference image conveying the target aesthetic, and a textual instruction that guides design intent and stylistic preferences. The ultimate goal is not only to generate photorealistic textures but also to adhere to strict user-defined spatial constraints while faithfully reflecting the specific aesthetic attributes captured within such reference images.


Despite this potential, achieving professional-grade panoramic interior generation remains a formidable challenge, primarily due to the intrinsic conflict between enforcing rigid structural constraints and incorporating flexible aesthetic guidance. Current frameworks often suffer from some problems. First, they frequently exhibit over-rigid conditioning that exhausts the ``generative budget'' on dense spatial layouts, leaving little room for hallucinating the rich decorative details essential for real-world designs. This over-dependency on dense geometric cues often results in under-decorated environments when the input layout is simplified. Second, when attempting to integrate heterogeneous visual inputs, a significant ``multi-condition interference'' arises. The latent structural cues embedded within a style reference image often directly conflict with the fixed architectural configuration of the place image. In practice, existing methods frequently prioritize stylistic patterns by implicitly relaxing geometric constraints, leading to undesired deformation of the original room configuration—such as shifted walls or misaligned layout boundaries—ultimately compromising structural fidelity in favor of stylistic consistency.

To address the intrinsic conflict between structural precision and aesthetic flexibility, we propose DreamHome-Pano, a controllable panoramic generation framework that effectively decouples geometric constraints from stylistic rendering. Current models often suffer from "over-rigid conditioning", where dense spatial layouts exhaust the generative budget, leading to visually plain environments. We resolve this by employing a dual-geometric prior: empty-room normal maps enforce rigid architectural boundaries while releasing the budget for decorative hallucination, and coarse instance segmentation provides flexible furniture guidance. To mitigate "multi-condition interference" caused by conflicting spatial signals in heterogeneous inputs, we standardize style references onto neutral spatial templates, isolating pure aesthetic attributes from native structural noise. In addition, we propose a Prompt-LLM to bridge the gap between structural layouts and aesthetic references. It synthesizes heterogeneous cues into professional instructions by completing missing material and color attributes for layout-defined elements not explicitly depicted in the reference, thereby ensuring global stylistic consistency. Furthermore, we establish a standardized panoramic evaluation benchmark covering 50 representative layouts and 10 styles to rigorously quantify design quality. Performance is assessed across three primary dimensions: spatial consistnecy is measured via pixel-wise mIoU for architectural and furniture alignment; aesthetic is evaluated using HPSv3 \cite{ma2025hpsv3}; realism is quantified through OmniAID \cite{guo2025omniaid}; and similarity is measured by CLIP score\cite{hessel2021clipscore}. To ensure the reliability of our assessment and capture professional nuances often missed by automated tools, we complement our analysis with a human expert evaluation led by interior design professionals. Finally, the framework achieves professional-grade performance through a hierarchical data-driven and progressive training paradigm, transitioning from foundational geometric alignment on a million-scale dataset to aesthetic refinement on expert-curated data.

Our main contributions are summarized as follows:

\begin{itemize}
    \item We propose DreamHome-Pano, a controllable panoramic generation framework that effectively decouples geometric constraints from stylistic rendering. Through a \textbf{Conflict-Free Control strategy} and a \textbf{Design-Aware Prompt-LLM}, the framework resolves the intrinsic tension between rigid structural layouts and flexible aesthetic guidance, ensuring high fidelity in both spatial configuration and design intent.
    \item We establish a standardized panoramic evaluation benchmark consisting of 50 representative layouts and 10 diverse design styles. This benchmark introduces a multi-dimensional assessment suite—covering spatial consistency, aesthetic, realism, and similarity —complemented by professional human expert evaluation.
    \item We implement a hierarchical and progressive training paradigm that transitions from million-scale foundational geometric alignment to expert-curated aesthetic refinement. This data-driven approach allows the model to achieve professional-grade performance, significantly outperforming existing methods in generating immersive, photorealistic, and structurally accurate 360-degree environments.
\end{itemize}

\section{Related work}

The emergence of Diffusion Models (DMs) has set a new benchmark for high-fidelity image synthesis. However, tailoring these models for professional interior design requires extending them from simple text-to-image generation to multi-condition, geometrically-constrained, and wide-angle panoramic synthesis. Our work builds upon two primary research directions: controllable image generation and panoramic image synthesis.

\subsection{Panoramic Image Generation}

Panoramic image generation has been widely explored to facilitate immersive 360-degree experiences. Most existing frameworks\cite{feng2023diffusion360,zhang2024taming,ni2025makes}, such as PanFusion and Diffusion360, are predominantly built upon text-to-image paradigms utilizing the UNet architecture. These methods primarily focus on ensuring boundary continuity in Equirectangular Projection (ERP) through modified attention mechanisms or circular convolutions.

However, several technical challenges persist in this field. From an architectural perspective, the reliance on conventional convolutional neural networks often limits the model's capacity for high-resolution texture synthesis and fine-grained detail compared to emerging transformer-based architectures. Regarding the training paradigm, current panoramic models are typically trained in a one-off supervised manner, often lacking hierarchical multi-stage refinement or human-preference alignment (e.g., RLHF). This absence of progressive optimization makes it difficult for models to consistently achieve professional-grade aesthetic quality and object plausibility.

\subsection{Controllable Image Generation}

Traditional text-to-image models often struggle with fine-grained spatial control. To address this, frameworks like ControlNet\cite{zhang2023adding} and IP-Adapter\cite{ye2023ip} introduced pluggable modules to incorporate structural priors such as Canny edges, depth maps, and segmentation masks.
However, these modular approaches primarily focus on single-condition injection and are inherently difficult to scale into an end-to-end, multi-image controlled pipeline.

More recently, a new generation of models has emerged that natively supports multi-image conditioning within a unified architecture. Models such as Qwen-Image-Edit\cite{wu2025qwen}, FLUX.2\cite{flux2024}, Seedream 4.5\cite{seedream2025seedream}, and Gemini 3 Pro Image\cite{google2025nanobananapro} leverage the scalability of Diffusion Transformers (DiTs) and multi-modal LLMs to process multiple visual prompts through token concatenation or dual-stream encoding. For instance, Seedream 4.5 unifies image composition and editing by pre-training on billions of text-image pairs, while FLUX.2 introduces a multi-reference feature that enables character and style consistency across up to ten images.

Despite their impressive generative capacity, current multi-modal models often suffer from over-rigid conditioning because their structural signals are directly extracted from ground-truth images during training, and they further struggle with signal entanglement between heterogeneous inputs due to the lack of training triplets that explicitly decouple structure, style, and target results.

\section{Data}

To facilitate high-fidelity panoramic generation, we developed a rigorous data curation pipeline that prioritizes both foundational image quality and professional aesthetic standards. Starting from an initial collection of approximately 2.55 million raw panoramic renderings, we implemented a hierarchical filtering and sampling strategy to derive our final training dataset. In addition, each curated image is paired with a professionally structured, attribute-grounded caption generated by a Vision-Language Model, providing fine-grained semantic supervision aligned with interior design principles.

\begin{figure}[H]
\centering
    \includegraphics[scale=0.5]{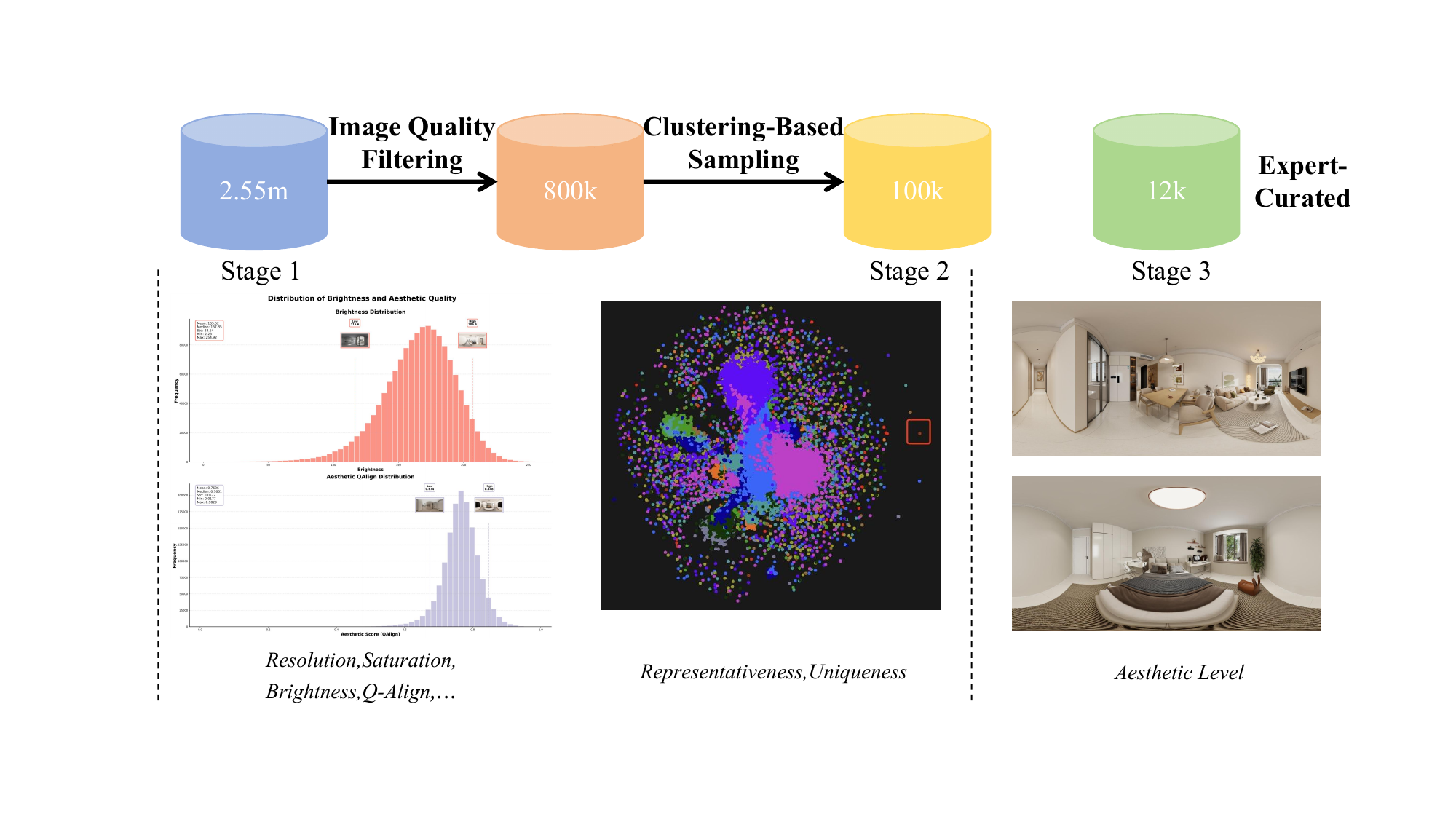}
    \caption{Hierarchical curation of high-quality panoramic interior images: quality filtering, clustering-based sampling, and expert aesthetic refinement.}
    \label{fig:data}
\end{figure}

\subsection{Hierarchical Curation Pipeline}

To achieve professional-grade panoramic generation, we implemented a three-stage hierarchical pipeline that progressively improves the model's structural fidelity and aesthetic quality. The pipeline transitions from large-scale foundational training at a lower resolution to high-resolution refinement on curated data.

\subsubsection{Stage 1: Massive Foundational Dataset}
\label{stage1 data}
The initial pool consists of $2.55m$ raw panoramic images collected from diverse interior design sources. At this stage, the data retains its original diversity but contains varying levels of quality. All samples in this stage are processed and stored at a resolution of $1024 \times 512$. This large-scale set provides the extensive visual patterns and geometric variety necessary for the model to understand basic panoramic projections.

\subsubsection{Stage 2: Quality-Filtered and Diversity-Sampled Dataset}
\label{stage2 data}
The second stage involves a rigorous refinement of the initial pool to ensure quality and diversity, resulting in a curated subset of $100k$ images.
\begin{itemize}
    \item Multi-Dimensional Filtering: The raw $2.55m$ images first undergo a quality assessment using heuristic filters (resolution, brightness, and contrast) and the Q-Align \cite{wu2023q} aesthetic scoring model, reducing the set to $800k$.
    \item Diversity Clustering: From the $800k$ candidates, we perform embedding-based clustering. We select $100k$ samples by balancing representativeness (proximity to cluster centers) and uniqueness (outlier detection) to eliminate repetitive patterns and ensure a wide coverage of interior archetypes.
    \item High-Resolution Standard: All images in this stage are upgraded to a resolution of $2048 \times 1024$, providing the necessary detail for high-quality synthesis.
\end{itemize}

\subsubsection{Stage 3: Expert-Curated Aesthetic Dataset}
\label{stage3 data}
The final stage focuses on peak aesthetic quality, comprising $12k$ ultra-high-standard images. Unlike the previous stages' automated filtering, this subset is manually curated by professional interior design experts. The selection focuses on professional-grade lighting, harmonious material textures, and superior composition.

\subsection{Hierarchical Attribute-Grounded Captioning}
\label{caption}
We utilize a Vision-Language Model (Seed 1.5-VL \cite{guo2025seed1}) to generate professional, attribute-dense descriptions through a structured two-step process guided by a hierarchical label system.

\begin{figure}[H]
\centering
    \includegraphics[width=1\linewidth]{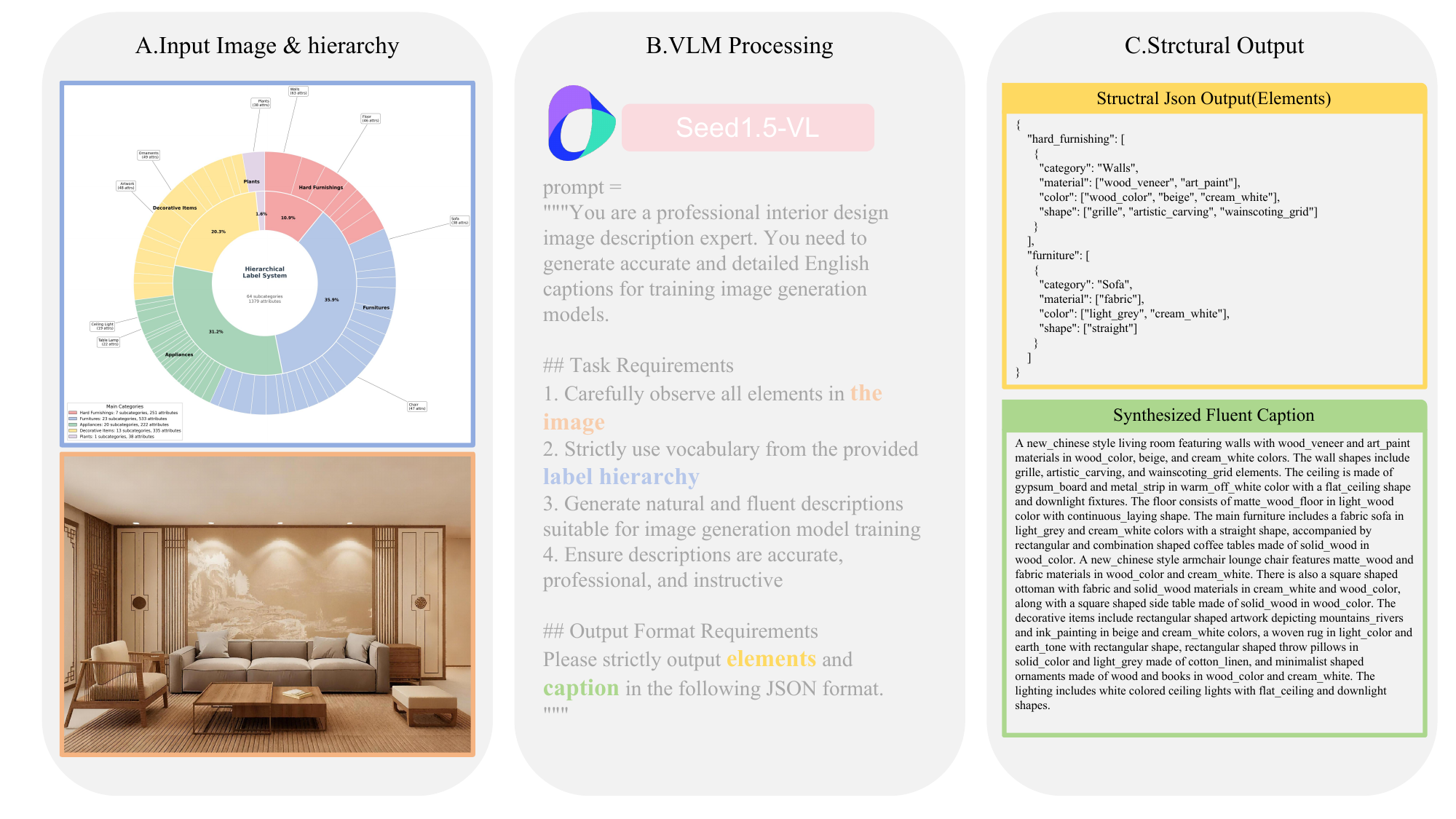}
    \caption{Hierarchical Attribute-Grounded Captioning Pipeline.}
    \label{fig:caption}
\end{figure}

\paragraph{Step 1: Element and Attribute Extraction}
The VLM identifies key interior elements (e.g., furniture, hard furnishings) and maps them directly to the 1,379 attributes in our hierarchy. This step ensures every item is described by specific materials, colors, and shapes.

\paragraph{Step 2: Fluent Caption Composition}
Based on the extracted attributes, the VLM synthesizes a single, comprehensive caption. This description integrates all identified design elements into a natural and fluent paragraph, providing high-fidelity semantic supervision for the generative model.

\section{Method}
\begin{figure}[htbp]
    \centering
    \includegraphics[width=1\linewidth]{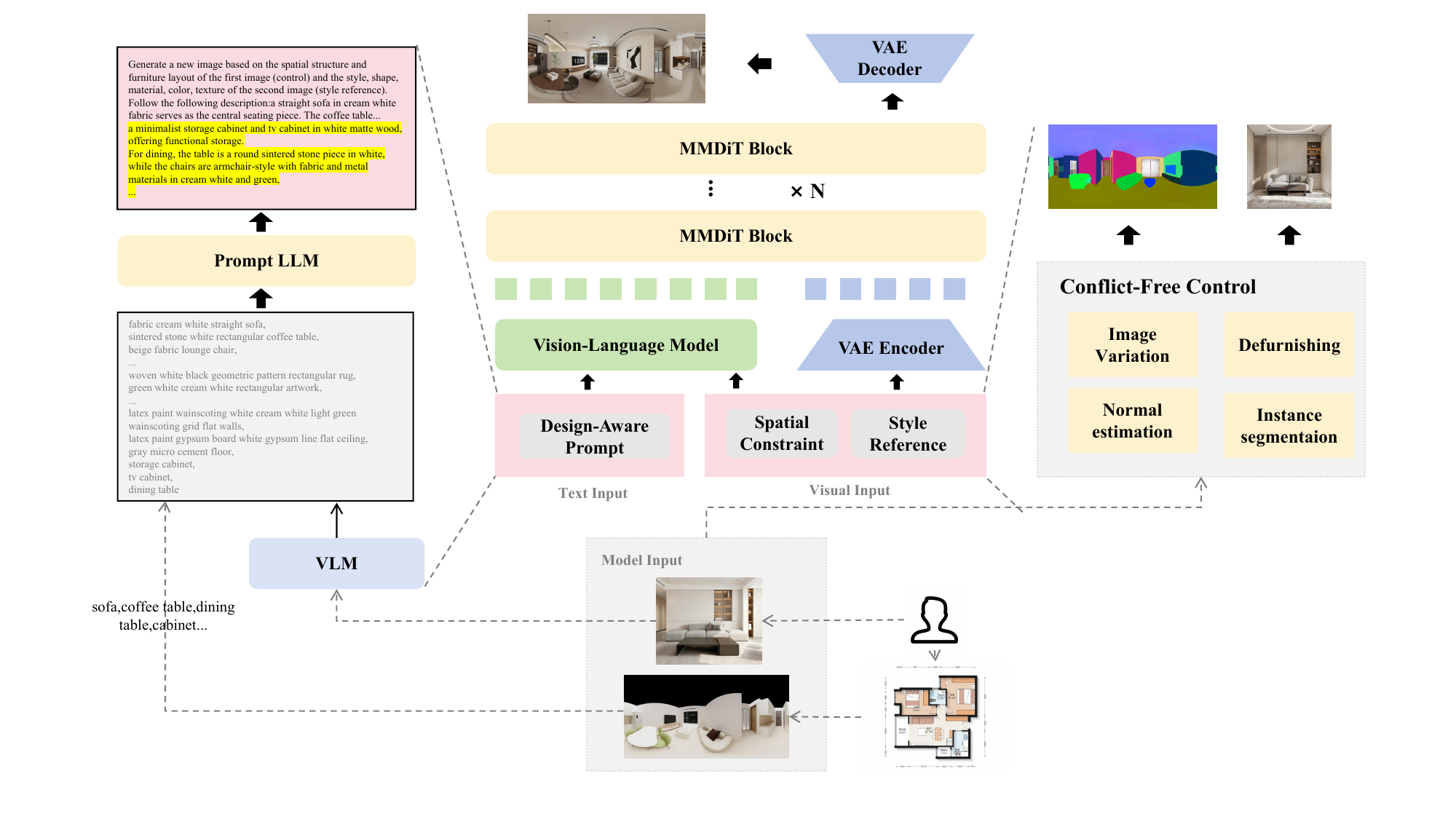}
    \caption{The Overall Architecture of DreamHome-Pano.}
    \label{fig:overview}
\end{figure}

DreamHome-Pano is a novel multi-modal conditional diffusion framework engineered for controllable panoramic interior generation. The framework maintains seamless compatibility with a broad spectrum of MM-DiT backbones, including Qwen-Image-Edit\cite{wu2025qwen} and FLUX.2\cite{flux-2-2025}. It comprises two key components: Prompt-LLM and Conflict-Free Control Strategy. As illustrated in Figure \ref{fig:overview}.

\subsection{Prompt-LLM}

We introduce a Prompt-LLM that generates a structurally consistent and stylistically coherent design description by integrating information extracted from both a spatial layout image and a style reference image. The Prompt-LLM serves as a design module that translates appearance cues into professional interior design semantics while strictly respecting layout constraints.

\subsubsection{Training Data and Style Prior}
The Prompt-LLM is trained on a curated corpus of panoramic interior scenes annotated with four complementary types of information.
Specifically, the \textit{style} $\mathcal{S}$ (e.g., French, Minimalism) and \textit{room type} $\mathcal{R}$ are manually annotated by professional interior designers to ensure authoritative and consistent design semantics.
In contrast, the \textit{interior elements} $\mathcal{E}$ and the corresponding \textit{detailed design descriptions} $\mathcal{D}$ are automatically generated using a Vision Language Model (VLM)~\cite{guo2025seed1}, following the hierarchical attribute-grounded captioning pipeline described in Section \ref{caption}.

Among them, only \textit{Decorative Items} and \textit{Plants} are treated as decorative elements, while all remaining categories correspond to core furnishings that define the functional and spatial structure of a room.

\subsubsection{Category-Aware Masking Strategy}
\begin{figure}[htbp]
    \centering
    \includegraphics[scale=0.5]{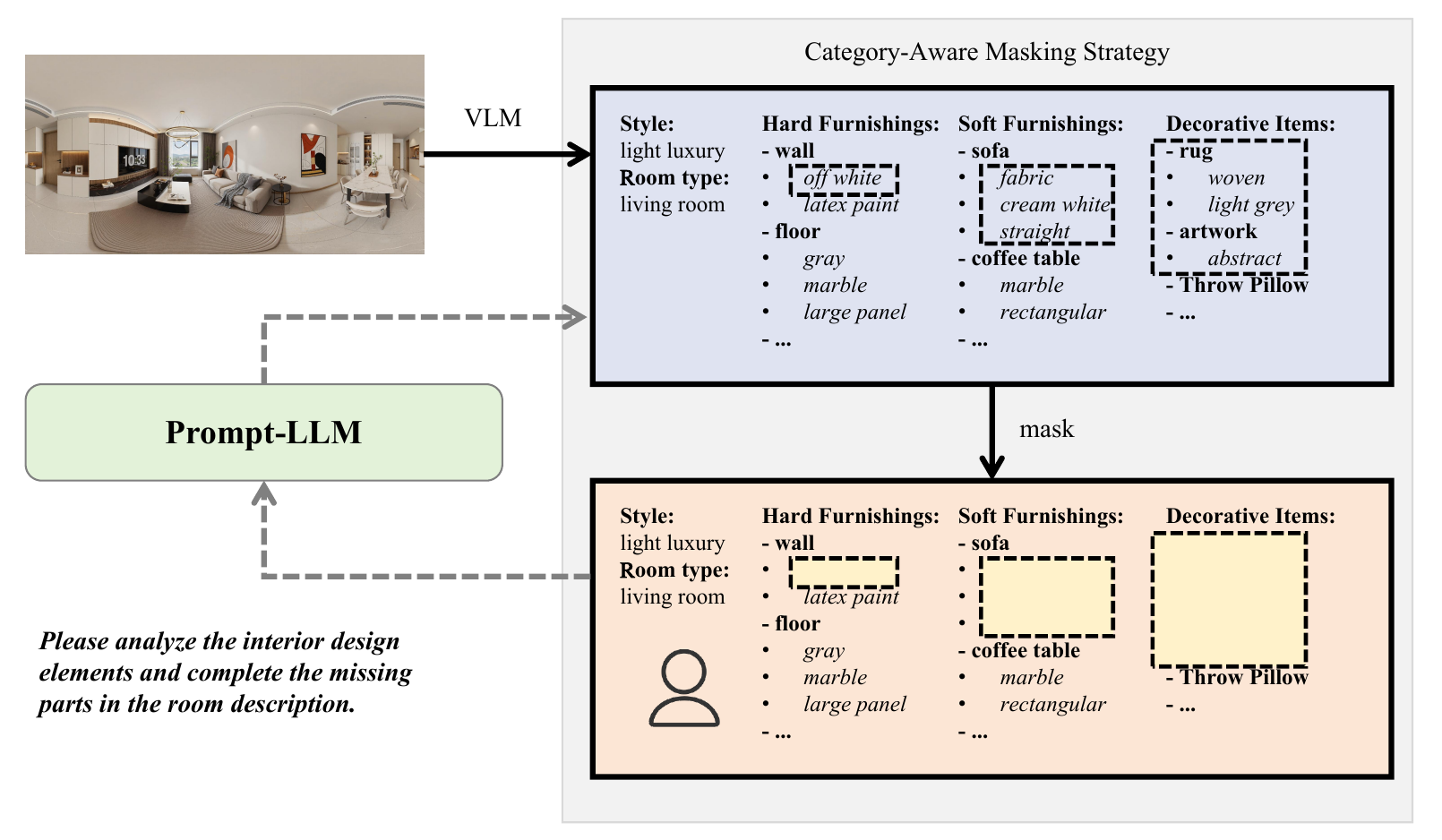}
    \caption{Training pipeline of Prompt-LLM featuring the Category-Aware Masking Strategy.}
    \label{fig:prompt-llm}
\end{figure}

We adopt a category-aware masking strategy over the element set $\mathcal{E}$. Specifically, interior elements are divided into core furnishings $\mathcal{E}_{\mathrm{fur}}$ and decorative elements $\mathcal{E}_{\mathrm{dec}}$, where
\begin{equation}
\mathcal{E}_{\mathrm{dec}} = \{\text{Decorative Items, Plants}\}.
\end{equation}
Each element is represented as a category–attribute pair $(c, a)$. During training, masking is applied asymmetrically: for core furnishings, category tokens are always preserved while only attribute tokens are randomly masked; for decorative elements, both category and attribute tokens are stochastically masked, As illustrated  in Figure \ref{fig:prompt-llm}.

Formally,
\begin{equation}
\tilde{\mathcal{E}} =
\mathcal{M}(\mathcal{E}) =
\big(
\{(c_i, m(a_i)) \mid (c_i,a_i)\in\mathcal{E}_{\mathrm{fur}}\},
\{(m(c_j), m(a_j)) \mid (c_j,a_j)\in\mathcal{E}_{\mathrm{dec}}\}
\big),
\end{equation}
where $m(\cdot)$ denotes a stochastic masking operator.

The Prompt-LLM is trained to predict the design description conditioned on masked inputs:
\begin{equation}
\mathcal{D} = f_{\text{Prompt-LLM}}(\mathcal{S}, \mathcal{R}, \tilde{\mathcal{E}}),
\end{equation}

This training design injects a strong inductive bias: the model learns to strictly respect the existence of core furnishings while acquiring the ability to complete missing attributes and generate stylistically appropriate decorative elements.

\subsubsection{Inference}
At inference time, given a user-specified style requirement, appearance attributes are extracted from the style reference image using Seed 1.5-VL \cite{guo2025seed1}.
Furniture categories not allowed by the place image are filtered out to prevent structural hallucination, while compatible attributes are transferred to the remaining categories. Conditioned on the place image constraints, transferred attributes, and the learned style prior, the Prompt-LLM completes missing furniture attributes and synthesizes decorative elements.
The resulting description $\mathcal{D}$ serves as a semantic anchor for the diffusion model, ensuring geometric consistency, stylistic fidelity, and professional design plausibility.

\subsection{Conflict-Free Control}

\begin{figure}[htbp]
    \centering
    \includegraphics[width=1\linewidth]{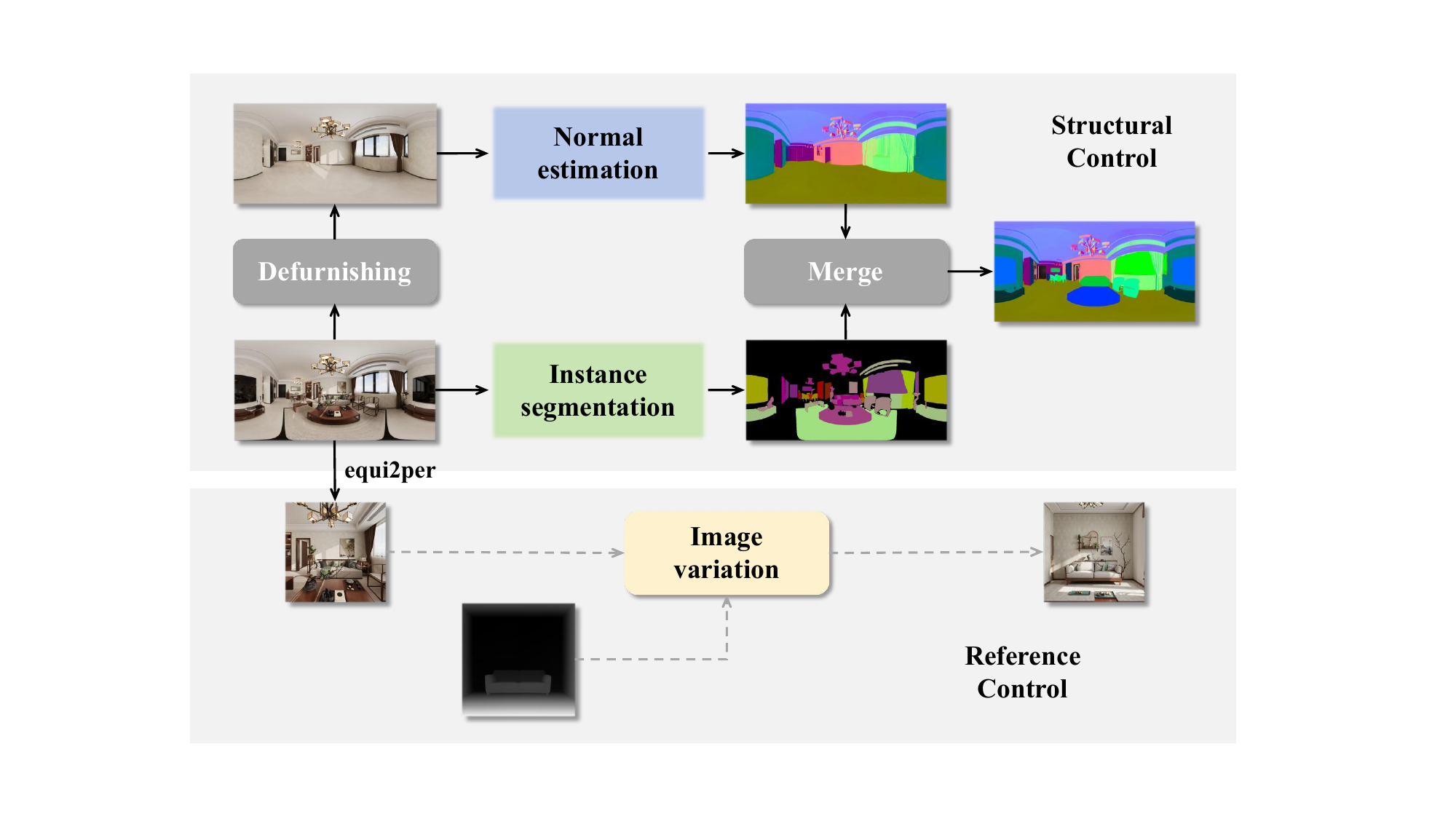}
    \caption{The \textbf{Conflict-Free Control Strategy}. The strategy decouples the control signal into two complementary parts: \textbf{Structural Control (Top)} and \textbf{Reference Control (Bottom)}. Structural Control fuses an empty-room Normal Map (rigid boundary constraint) with a furniture Instance Segmentation Map (flexible interior guidance) to create a unified geometric prior. Reference Control provides a style signal that is free from the reference image’s native structure, effectively mitigating the structural-aesthetic conflict. }
    \label{fig:conflict-Free Control Framework}
\end{figure}

The primary challenge in conditional panorama generation lies in the conditioning trade-off, wherein enforcing geometric constraints can conflict with the application of flexible aesthetic styles. To address this challenge, we propose the \textbf{Conflict-Free Control Strategy}, designed to decouple structural and aesthetic signals. As illustrated in Figure \ref{fig:conflict-Free Control Framework}, the strategy operates through two distinct yet complementary modules: Structural Control and Reference Control. \textbf{Structural Control} establishes the fundamental structural constraints using a dual geometric prior strategy, thereby ensuring spatial coherence. \textbf{Reference Control} isolates the aesthetic style from the reference image’s native spatial structure, ensuring that the style guidance does not violate the prescribed structural constraints. This principled decoupling is critical for generating high-fidelity panoramic interiors that simultaneously satisfy both geometric consistency and aesthetic preferences. 

\subsubsection{\textbf{Structural Control}}

Structural Control constitutes the core module of DreamHome-Pano’s dual geometric prior strategy, tasked with enforcing strict structural constraints while maintaining creative flexibility for interior elements. Its workflow comprises two principal steps: empty-room normal estimation and furniture-instance segmentation, which are subsequently integrated to produce a unified structural control signal.

\begin{enumerate}
    \item \textbf{Empty-Room Normal Estimation:} We first derive the geometric prior of the room’s fixed structure via an empty-room normal map. 
    \begin{itemize}
        \item \textbf{Defurnishing Preprocess:} Beginning with a furnished target panorama, we employ a defurnishing module (Gemini 3 Pro Image\cite{google2025nanobananapro} and a fine-tuned Qwen-Image\cite{wu2025qwen}) to remove all interior elements (e.g., sofa, table). The resulting empty-room panorama preserves the room’s inherent architectural structure, including walls, ceilings, and windows. 
        \item \textbf{Normal Estimation:} A panoramic normal estimation model, Moge-2\cite{wang2025moge}, is applied to the empty-room panorama to generate a normal map. This map encodes precise, high-frequency geometric cues at critical structural junctions (e.g., wall-ceiling corners, window frames) and serves as a rigid constraint, ensuring that the generated panorama strictly conforms to the room’s boundaries and structural consistency.
    \end{itemize}

    \item \textbf{Coarse Instance Segmentation:} To balance layout rigidity with decorative flexibility, we introduce a coarse semantic segmentation map for furniture to fix the layout of interior elements. 
    \begin{itemize}
        \item \textbf{Instance Segmentation:} The furnished target panorama is processed using a custom instance segmentation model—an integration of our detection model and the SAM model\cite{kirillov2023segment}—to generate a mask map that labels the precise category and spatial location of core furniture (e.g., sofas, coffee tables) . 
        \item \textbf{Flexible Constraints:} In contrast to the strict boundaries provided by the normal map, this segmentation map functions as a flexible guide, regulating the approximate placement and type of furniture rather than exact shapes, thereby preserving space for aesthetic customization.
    \end{itemize}

    \item\textbf{Merged Structural Control Signal:} We fuse the empty-room normal map and coarse instance segmentation map into a unified control tensor. This merged signal encodes two complementary constraints:
        \begin{itemize}
            \item Rigid structural constraints derived from the normal map, which preserve architectural consistency;
            \item Flexible furniture guidance derived from the segmentation map, which aligns interior elements with the intended design. 
        \end{itemize}
        By decoupling and integrating these two geometric priors, the Structural Control module ensures the generated panorama maintains both spatial coherence (avoiding distorted walls/furniture) and creative freedom (supporting diverse stylistic configurations of interior elements). 
\end{enumerate}

\subsubsection{\textbf{Reference Control}}

The Reference Control module addresses the challenge of style transfer without compromising spatial structure by normalizing reference images to a unified spatial template—a design decision that facilitates the disentanglement of stylistic attributes from the native structure of reference image.
\begin{itemize}
    \item \textbf{Template Depth Map Definition:} We predefine a domain-specific standard depth map (e.g., a standard bedroom depth template) that encodes fixed coordinates for structural elements (walls, windows) and furniture placement ranges. This ensures that all reference images provided to the model share a consistent spatial structure derived from the same template. 
    \item \textbf{Spatial Transformation and Normalization:} We leverage FLUX-Depth \cite{flux2024} as a spatial transformer to decouple the reference image into two distinct components: its native spatial layout and its aesthetic appearance. By taking both the arbitrary style reference and the standard depth template as inputs, the model warps the reference content to conform to the template's predefined spatial structure. This process explicitly discards the original spatial information while strictly preserving the appearance attributes (e.g., color palettes, material textures, and decor patterns). The resulting "\textbf{standard reference image}" neutralizes the original spatial correspondence between the reference and the target result, effectively reducing the variance of spatial configurations in the style inputs. Consequently, this strategy ensures that the reference image serves as a pure stylistic signal without imposing its native structure on the generation process, allowing the model to focus exclusively on its aesthetic attributes.
\end{itemize}

\section{Training}
The training process of DreamHome-Pano is structured into two main stages: Supervised Fine-Tuning (SFT) and Reinforcement Learning (RL).

\subsection{Supervised Fine-Tuning (SFT)}
The SFT process is conducted entirely within the latent space of a pre-trained Variational Autoencoder (VAE) to ensure computational efficiency and high-fidelity reconstruction. To minimize the interference of style reference images on the underlying structural layout, we introduce a Latent Mask Strategy. For a given style reference latent $z_{ref}$, we apply a stochastic binary mask $M$, yielding $\hat{z}_{ref} = M \odot z_{ref}$. This operation is designed to suppress spatial leakage by disrupting the pixel-to-pixel correspondence between the reference image and the target panorama. By perturbing these spatial cues, we force the model to transcend simple low-level copying and instead learn abstract stylistic representations, such as color palettes, material palettes, and lighting atmospheres.

The training follows a structured three-stage curriculum, progressively transitioning from coarse geometric alignment to fine-grained aesthetic refinement.

\subsubsection{Stage 1: Large-scale Alignment and Panoramic Distribution Calibration.}

The primary objective of this initial stage is two-fold: establishing a rigorous mapping between structural control signals and panoramic geometry, and calibrating the generative distribution of the base model. This stage is conducted at a standard resolution (e.g., 512$\times$1024), which allows the model to efficiently learn global spatial relationships and panoramic projection priors.  

Since the backbone model is pre-trained predominantly on standard perspective images, its generative prior inherently lacks an understanding of the equirectangular projection used in 360-degree panoramas. To address this, we leverage a million-scale panoramic dataset to perform domain adaptation, as detailed in Section \ref{stage1 data}. This process forces the model to correct its generative distribution, learning to synthesize the panoramic distortions and ensuring seamless horizontal continuity across the 360-degree field of view. Simultaneously, to ensure the model prioritizes structural integrity over stylistic interference, we implement a gradual learning schedule for the multi-task mixing ratio. Training commences with a 10:0 ratio, focusing exclusively on "structural-only" tasks, and gradually transitions to a stable 2:8 ratio of single-condition to multi-condition tasks. This "geometry-first" strategy ensures the model masters the fundamental structural integrity and panoramic projection before being introduced to complex stylistic constraints.

\subsubsection{Stage 2: High-resolution Fine-tuning and Texture Refinement.} 

Building upon the geometric foundation established in Stage 1, the second stage involves scaling up the training resolution (e.g., to 1024$\times$2048) to enhance the model's ability to render high-frequency details and intricate textures. As described in Section \ref{stage2 data}, we utilize a curated dataset of $100k$ high-quality panoramic images specifically selected for their visual quality.

During this stage, the multi-task mixing ratio is fixed at 2:8 to maintain a balanced learning signal between layout adherence and style transfer. Transitioning to a higher resolution at this stage prevents the model from being overwhelmed by pixel-level noise during early geometric learning, ensuring that the refined textures are anchored to a stable and accurate architectural structure.

\subsubsection{Stage 3: Aesthetic Enhancement and Design Harmony.} 

The final stage of SFT is dedicated to elevating the visual output from technical photorealism to professional interior design standards. For this stage, we utilize a highly selective subset of $12k$ panoramas. As detailed in Section \ref{stage3 data}, these samples represent the "gold standard" of our dataset, characterized by expert-level lighting compositions, sophisticated color palettes, and harmonious furniture arrangements.

During this stage, the training objective shifts from rendering small details to improving the overall lighting and color balance. By the end of this stage, DreamHome-Pano can generate interior spaces that not only follow the structural constraints perfectly but also look like professional photographs with natural lighting and a comfortable atmosphere.

\subsection{Reinforcement Learning (RL)}
To further align the model with professional design preferences and resolve subtle artifacts, we perform RL optimization in two phases:

\subsubsection{Phase 1: Direct Preference Optimization (DPO).} We construct a self-contrast preference dataset specifically for interior design. The ground-truth (GT) training images are designated as the \textit{chosen} (positive) samples, while the model's own inference results—which might exhibit minor structural distortions or blurred textures—are used as the \textit{rejected} (negative) samples. This enables the model to learn a clear boundary between professional-grade renders and common generation artifacts.
    
\subsubsection{Phase 2: Reward-driven Fine-tuning via Diffusion-NFT}
In the final refinement stage, we adopt the \textbf{Diffusion-NFT (Non-adversarial Fine-Tuning)}\cite{zheng2025diffusionnft} framework to optimize the generation policy within a multi-objective reward space. Unlike traditional reinforcement learning methods that often suffer from training instability, Diffusion-NFT treats the reward-guided fine-tuning as a specialized sampling process. This allows us to directly backpropagate gradients from a composite reward function $R$ into the diffusion model, ensuring stable convergence even with complex, high-dimensional panoramic outputs. The reward function $R$ evaluates the model across three critical dimensions:

\begin{itemize}
    \item \textbf{Structural Fidelity Reward:} This metric utilizes a pre-trained layout detection model to calculate the Intersection over Union (IoU) between the generated furniture/architectural components and the input instance masks. It ensures that the model maintains strict adherence to the specified layout.
    \item \textbf{Architectural Plausibility Reward (OmniAID):} We leverage the OmniAID metric to provide a high-level realism score. This reward evaluates whether the generated room maintains a physically plausible structure and a realistic distribution of interior elements.
    \item \textbf{Semantic Intent Reward (LongCLIP):} To ensure the model faithfully follows complex, design-heavy text prompts, we use LongCLIP to measure the semantic alignment. This reward encourages the model to accurately render specific design motifs and material descriptions requested by the user.
    \item \textbf{Aesthetic Preference Reward (HPSv3):} We incorporate the Human Preference Score v3 (HPSv3) to align the model with general human aesthetics. This reward focuses on improving the overall visual quality, such as better lighting, cleaner textures, and more appealing color compositions, making the final results look more professional and polished.
\end{itemize}

\section{Experiments}

\begin{figure}[htbp]
    \centering
    \includegraphics[width=1\linewidth]{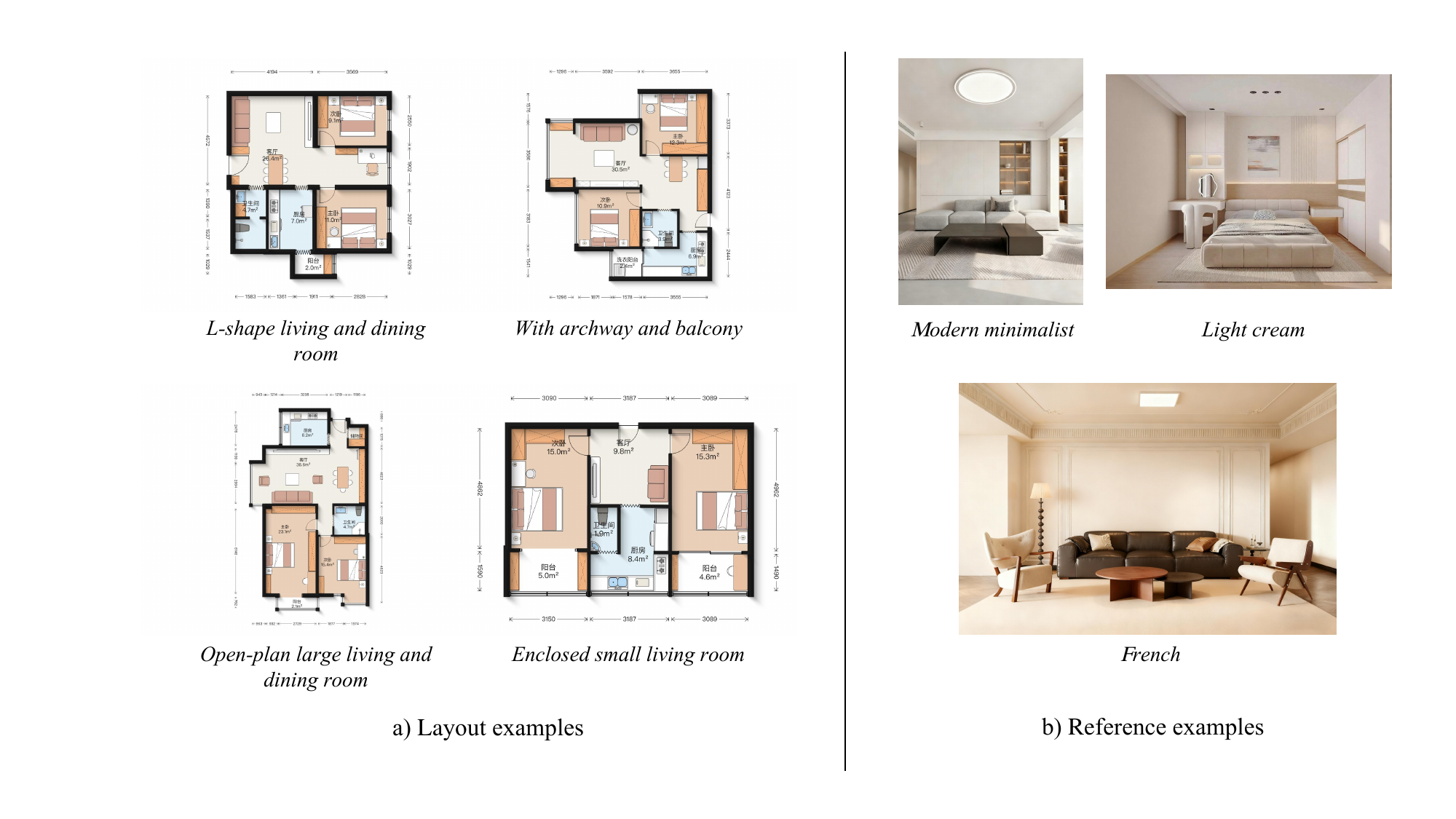}
    \caption{Evaluation Data Examples}
    \label{Evaluation Data Examples}
\end{figure}

\subsection{Evaluation Data}

To evaluate the structural fidelity and style transfer performance of DreamHome-Pano, we constructed a structured test set organized along two primary dimensions: floorplan layout (structural constraints) and design style (aesthetic guidance), as shown in Figure \ref{Evaluation Data Examples}.

\begin{itemize}
    \item \textbf{Floorplan Layouts:} We selected $\mathbf{50}$ representative layouts covering diverse architectural structures and complexities. These serve as fixed geometric priors to verify the model's ability to preserve structural integrity, ensuring a diverse range of spatial challenges.

    \item \textbf{Style Reference Sets:} We chose $\mathbf{10}$ mainstream and high-aesthetic design style sets. Each set contains multiple reference images corresponding to different functional spaces (e.g., living room, bedroom, kitchen). These style references cover a wide range of popular aesthetics and are used to evaluate the model's effectiveness in transferring aesthetic attributes across different style domains.
\end{itemize}

\subsection{Evaluation Protocol}

\subsubsection{Automated Machine Evaluation}
\begin{figure}
    \centering
    \includegraphics[width=1\linewidth]{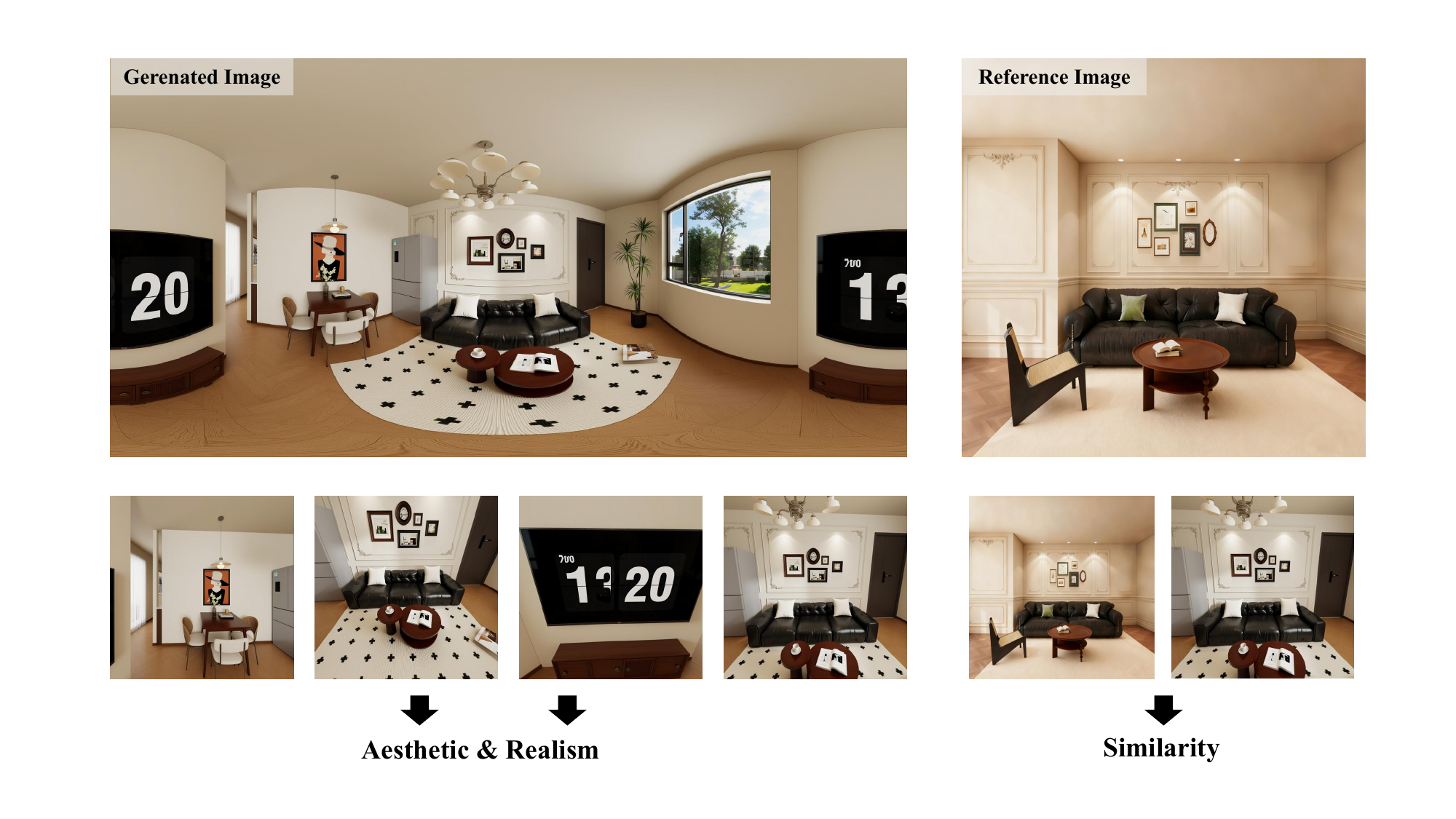}
    \caption{Illustration of the content-aware evaluation process.}
    \label{fig:perspective eval}
\end{figure}

To ensure a comprehensive and unbiased evaluation, we employ a dual-domain measurement strategy. Spatial Consistency is computed directly on the equirectangular panoramas to capture the global geometric integrity and 360° boundary continuity. In contrast, to evaluate Aesthetic Harmony (HPSv3), Reference Similarity (CLIP), and Visual Plausibility (OmniAID), we project each panorama into a series of Normal Field-of-View (NFoV) perspective views. As illustrated in \ref{fig:perspective eval}, these projections are strategically centered on the spatial coordinates of core furniture instances to ensure the perceptual metrics focus on the most semantically significant regions of the interior design.  This approach avoids the inherent geometric distortion of equirectangular projections, ensuring that the pre-trained perceptual models—which are optimized for standard photographic domains—can accurately assess fine-grained textures, lighting, and semantic realism as perceived by a human observer. 
\begin{itemize}
    \item \textbf{Spatial Consistency:} This criterion are designed to identify hard failures\ in geometric fidelity by quantifying the pixel-level alignment between the input constraints (i.e., the place image) and the generated panoramas. Specifically, we employ a layout estimation model \cite{jiang2022lgt} to extract the room layout (e.g., walls, ceilings, and floors), and a semantic segmentation model \cite{cheng2022masked} to obtain pixel-wise masks for all remaining semantic elements, including doors, windows, and core furniture. The generated outputs are then compared against the corresponding masks derived from the place images using pixel-wise Intersection-over-Union (IoU). This unified evaluation protocol enforces strict adherence to the prescribed spatial layout while providing a consistent and interpretable measure of overall geometric correctness.
    \item \textbf{Aesthetic:} To assess overall visual appeal, we adopt HPSv3 \cite{ma2025hpsv3}, a human-preference-aligned scoring model that has been shown to correlate well with human judgments of image aesthetics. Higher HPSv3 scores indicate stronger perceptual quality and design plausibility.
    \item \textbf{Realism:} We utilize the OmniAID score \cite{guo2025omniaid} to evaluate physical and semantic integrity. Its decoupled MoE architecture independently assesses high-level structural logic and low-level generative artifacts, where higher scores signify superior photorealism and professional-grade consistency in synthesized 360° scenes. 
    \item     \textbf{Similarity:} We employ both CLIP score \cite{hessel2021clipscore}. CLIP score measures cross-modal semantic consistency by computing the cosine similarity between image embeddings in a vision–language joint space, capturing high-level stylistic and semantic correspondence.
\end{itemize}

\subsubsection{Expert Human Evaluation}

To capture the professional nuances that may not be adequately reflected by automated metrics, we convened a panel of interior design experts to perform a comprehensive evaluation. The expert evaluation was guided by three key dimensions:

\begin{itemize}
    \item \textbf{Spatial Consistency (30\%):} This criterion evaluates both the geometric stability of the 360-degree panorama and its adherence to the input structural constraints. Experts verify the absence of projection-related distortions (e.g., curved or warped walls) and assess whether the generated room structure and furniture arrangement faithfully correspond to the prescribed architectural specifications.
    \item \textbf{Plausibility (30\%):} This dimension examines the realism of individual generated objects, focusing on whether textures, scales, and material properties (e.g., marble reflections or wood grain patterns) are visually and structurally credible, and free from geometric artifacts.
    \item \textbf{Aesthetic (40\%):} This criterion assesses the overall design coherence, including color harmony, lighting consistency, and conformity to the intended style as specified by the design-aware prompt.
\end{itemize}

Based on these dimensions, a comprehensive score ($S_{total}$) is calculated as a weighted sum of the individual scores (each on a scale of 1 to 5):

\begin{equation}
S_{total} = (S_{aesthetic} \times 0.4) + (S_{spatial\ consistency} \times 0.3) + (S_{plausibility} \times 0.3)
\end{equation}

Each generated panorama is then classified into one of five quality tiers according to its final score:

\begin{itemize}
    \item \textbf{Grade S (Excellent, $S_{total} \geq 4.0$):} Results exhibit professional-grade quality with flawless spatial logic and exceptional aesthetic appeal.
    \item \textbf{Grade A (High Quality, $3.5 \leq S_{total} < 4.0$):} Results are highly usable, demonstrating strong structural integrity and minor, negligible aesthetic imperfections.
    \item \textbf{Grade B (Acceptable, $2.5 \leq S_{total} < 3.5$):} Results show average performance, with basic adherence to layout but limited detail refinement.
    \item \textbf{Grade C (Marginal, $1.0 < S_{total} < 2.5$):} Results suffer from noticeable structural or stylistic inconsistencies, requiring significant manual intervention.
    \item \textbf{Grade D (Unusable, $S_{total} \leq 1.0$):} Results exhibit severe geometric collapse or aesthetic failure, failing to meet basic design requirements.
\end{itemize}

\subsection{Main Results}

\subsubsection{Quantitative evaluation }

To comprehensively evaluate the performance of DreamHome-Pano, we employ a dual-track evaluation consisting of Automated Machine Metrics and Expert-based Human Evaluation. This combination allows us to capture both the rigid structural fidelity and the nuanced aesthetic quality of the generated 360-degree scenes.

Furthermore, to demonstrate the architecture-agnostic versatility of our framework, we implement DreamHome-Pano on two distinct backbones: \textbf{Qwen-Image-Edit}\cite{wu2025qwen} and \textbf{Flux.2-dev}\cite{flux-2-2025}, while the Flux-based version is evaluated in its post-SFT state to showcase the framework’s strong foundational performance even without specialized RL optimization.
Additionally, we conduct qualitative and quantitative comparisons against state-of-the-art closed-source models (Seedream 4.5\cite{seedream2025seedream} and  Gemini 3 Pro Image\cite{google2025nanobananapro}). 

\begin{table}[H]
    \centering
    \footnotesize  
    \caption{Automated Machine Metrics.}
    \label{tab:iou quantitative evaluation}
    \setlength{\tabcolsep}{1.5pt}  
    \renewcommand{\arraystretch}{1.1}  
    \begin{tabular}{ccccccccccc}
        \toprule
        & \multicolumn{7}{c}{\textbf{Spatial Consistency}} & \textbf{Aesthetic}& \textbf{Similarity}& \textbf{Realism}\\
        \cmidrule(lr){2-8} \cmidrule(lr){9-9} \cmidrule(lr){10-10} \cmidrule(lr){11-11}
        & Wall & Door & Window & Cabinet & Sofa & Bed & Average & HPSv3 & CLIP score & OmniAID score\\\midrule
 Seedream 4.5& 0.8028& 0.2769& 0.2000& 0.2510& 0.3336& 0.5155& 0.3966& 4.3410& 0.8491&0.6746\\ 
 Gemini 3 Pro Image& 0.8916& 0.4379& 0.3453& 0.5361& 0.3500& 0.6099& 0.5285& 6.9970& 0.8255&0.2587\\
 Ours (FLUX.2)& 0.9693& 0.6578& 0.4770& 0.6816& 0.6956& 0.7423& 0.7039& 6.4672& 0.8182&0.2233\\ 
 \textbf{Ours (Qwen-Image-Edit)}& 0.9650& 0.6770& 0.5416& 0.5594& 0.7296& 0.7489& 0.7036& 7.4648& 0.8230&0.1793\\ \bottomrule 
    \end{tabular}
\end{table}

Table \ref{tab:iou quantitative evaluation}  provides a comprehensive quantitative comparison between our framework and current state-of-the-art baselines, highlighting that our models—particularly the Qwen-based variant—demonstrate a decisive advantage in balancing structural fidelity with aesthetic synthesis. Both of our variants excel in rendering the architectural "shell" with high mIoU, effectively eliminating the "geometric drift" and structural hallucinations  prevalent in Seeddream 4.5 and Gemini 3 Pro Image. Furthermore, there is an expected trade-off in raw OmniAID scores as our models prioritize complex architectural logic and professional design over basic pixel-level smoothness, the significant lead in spatial harmony and aesthetic quality proves that our framework successfully elevates AI generation to a professional architectural standard.

\begin{table}[H]
    \centering
    \caption{Expert-based Human Evaluation.}
    \label{tab:expert human evaluation}
    \setlength{\tabcolsep}{1.5pt}  
    \renewcommand{\arraystretch}{1.1}  
    \begin{tabular}{cccccccccc}
        \toprule
        & \multicolumn{4}{c}{\textbf{Score}} & \multicolumn{5}{c}{\textbf{Grade}}\\
        \cmidrule(lr){2-5} \cmidrule(lr){6-10}
                      & Spatial Consistency& Aesthetic& Plausibility& Total & S & A & B & C& D\\ \midrule
      Seedream 4.5    & 1.66& 1.26& 1.80& 1.54& 1\%& 12\%& 8\%& 1\%&78\%\\ 
 Gemini 3 Pro Image   & 3.33& 2.66& 3.79& 3.20& 27\%& 38\%& 13\%& 0\%&22\%\\
    Ours (FLUX.2)     & 4.12& 3.21& 3.85& 3.68& 40\%& 37\%& 16\%& 0\%&6\%\\ 
\textbf{Ours (Qwen-Image-Edit})& 4.28& 3.20& 3.91& 3.74& 58\%& 26\%& 5\%& 0\%&11\%\\ \bottomrule 
    \end{tabular}
\end{table}
The results summarized in Table \ref{tab:expert human evaluation} underscore the distinct advantages of our framework, particularly the Qwen-based variant, in meeting professional interior design standards. While automated metrics provide a baseline, the expert evaluation reveals a more nuanced picture of how our models bridge the gap between "generative plausibility" and "professional utility."

First, the Qwen-based variant emerges as the superior configuration for high-end applications, achieving the highest concentration of Grade S (Excellent) results at 58\%. This is a significant margin over both the FLUX-based variant (40\%) and the competitive Gemini 3 Pro Image (27\%). This concentration of top-tier grades indicates that when guided by the Qwen-based backbone, the framework is exceptionally capable of producing "professional-grade" outputs where spatial logic, lighting harmony, and material authenticity align perfectly.

Second, in terms of Spatial Consistency, the Qwen-based variant achieves a peak score of 4.28, the highest in the study. This score reflects its robust ability to adhere to architectural specifications without the "warped wall" artifacts or projection distortions that plague baselines like Seedream 4.5 (1.66). The high Plausibility score (3.91) further confirms that the model synthesizes textures and furniture scales that are structurally credible to the trained eye of a designer.

Finally, while the FLUX-based variant demonstrates a slightly more stable "floor" with fewer Grade D failures (6\% vs. 11\%), the Qwen-based model represents the true ceiling of the framework's capabilities. With a Total Score of 3.74, it effectively balances the rigorous demands of spatial consistency (30\%) and aesthetic (40\%). These expert insights validate that our DreamHome-Pano framework, particularly when instantiated with the Qwen-Image-Edit backbone, successfully elevates panoramic generation to a level of professional-grade architectural visualization.

\subsubsection{Qualitative Evaluation}
\begin{figure}[H]
     \centering
     \includegraphics[width=1\linewidth]{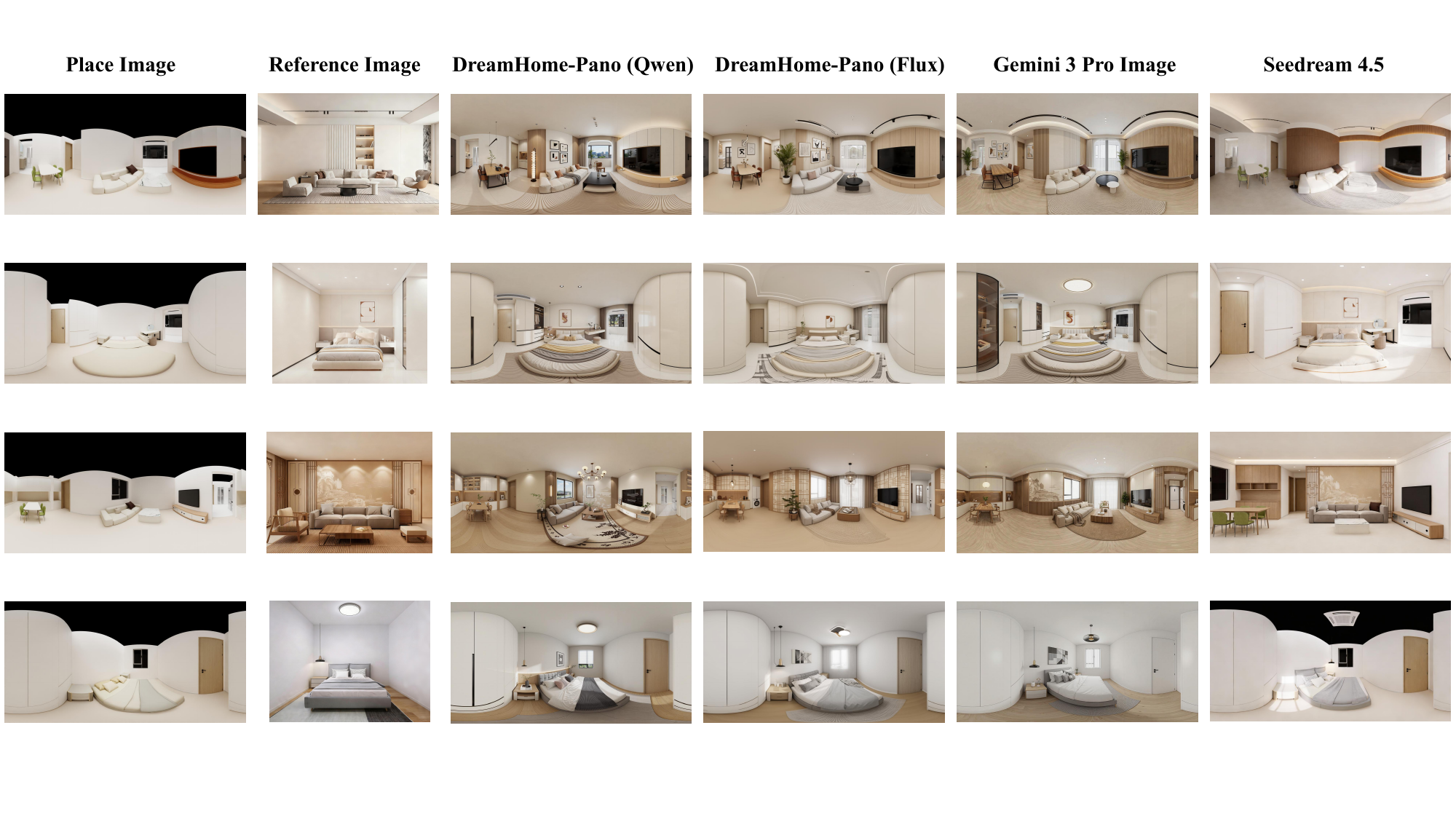}
     \caption{Qualitative comparison of panoramic interior generation results between DreamHome-Pano and state-of-the-art baselines (Seedream 4.5 and Gemini 3 Pro Image) under consistent structure and style constraints. }
     \label{fig:Qualitative vs nano/seed}
 \end{figure}

As illustrated in Figure \ref{fig:Qualitative vs nano/seed}, DreamHome-Pano exhibits clear advantages over existing baselines in terms of both structural fidelity and aesthetic quality. Seedream 4.5\cite{seedream2025seedream} frequently suffers from severe geometric distortions and implausible lighting conditions, leading to results that deviate substantially from professional interior design standards. Gemini 3 Pro Image\cite{google2025nanobananapro}, while capable of preserving coarse spatial consistency, often struggles with fine-grained detail control and material rendering, resulting in blurred boundaries and inconsistent surface textures. In contrast, our framework consistently produces photorealistic and structurally coherent panoramas. By jointly leveraging the Design-Aware Prompt-LLM and the Conflict-Free Control strategy, DreamHome-Pano generates refined visual details—such as sharp object boundaries and physically plausible material reflections—while strictly conforming to the prescribed layout constraints and style references, yielding results comparable to professional architectural photography.

\newpage  
\subsection{Ablation Study}

\subsubsection{Three-Stage Supervised Fine-Tuning}

\begin{table}[H]
    \centering
    \footnotesize  
    \caption{Quantitative ablation study of three-stage supervised fine-tuning. }
    \label{tab:sft metrics}
    \setlength{\tabcolsep}{1.5pt}  
    \renewcommand{\arraystretch}{1.1}  
    \begin{tabular}{ccccccccccc}
        \toprule
        & \multicolumn{7}{c}{\textbf{Spatial Consistency}} & \textbf{Aesthetic}& \textbf{Similarity}& \textbf{Plausibility}\\
        \cmidrule(lr){2-8} \cmidrule(lr){9-9} \cmidrule(lr){10-10} \cmidrule(lr){11-11}
        & Wall & Door & Window & Cabinet & Sofa & Bed & Average & HPSv3 & CLIP score & OmniAID score\\\midrule
 \makecell{Base + Stage 2}& 0.9636& 0.6071& 0.6067& 0.5398& 0.6623& 0.7388& 0.6760& 7.3395& 0.8218&0.1763\\ 
        
 \makecell{Stage 1 + Stage 2}& 0.9648& 0.6174& 0.5392& 0.5009& 0.7153& 0.7596& 0.6829& 6.3684& 0.8184&0.1836\\
 \textbf{\makecell{Stage 1 + Stage 2 + Stage 3}}& 0.9629& 0.6699& 0.5275& 0.5417& 0.7295& 0.7486& 0.6967& 7.1441& 0.8251&0.1766\\ \bottomrule 
    \end{tabular}
\end{table}

To ablate the contribution of each individual training stage, we conducted detailed ablation experiments based on the Qwen-Image-Edit backbone. As summarized in Table \ref{tab:sft metrics}, incorporating the third fine-tuning stage yields the most balanced overall performance across spatial fidelity, aesthetic quality, and reference alignment. The full three-stage model (Stage 1 + Stage 2 + Stage 3) achieves the highest average spatial consistency (0.6967), indicating improved robustness in preserving both architectural structures and furniture layouts. Moreover, it attains the best CLIP score (0.8251), reflecting stronger semantic alignment with the reference image, while maintaining competitive aesthetic quality as measured by HPSv3. Compared to partial fine-tuning variants, these results demonstrate that the third stage effectively consolidates geometric supervision and style-aware refinement, leading to a more harmonious trade-off between structural accuracy, visual plausibility, and stylistic consistency. 
\begin{figure}[H]
    \centering
    \begin{subfigure}[b]{0.48\textwidth}
        \centering
        \includegraphics[width=\textwidth]{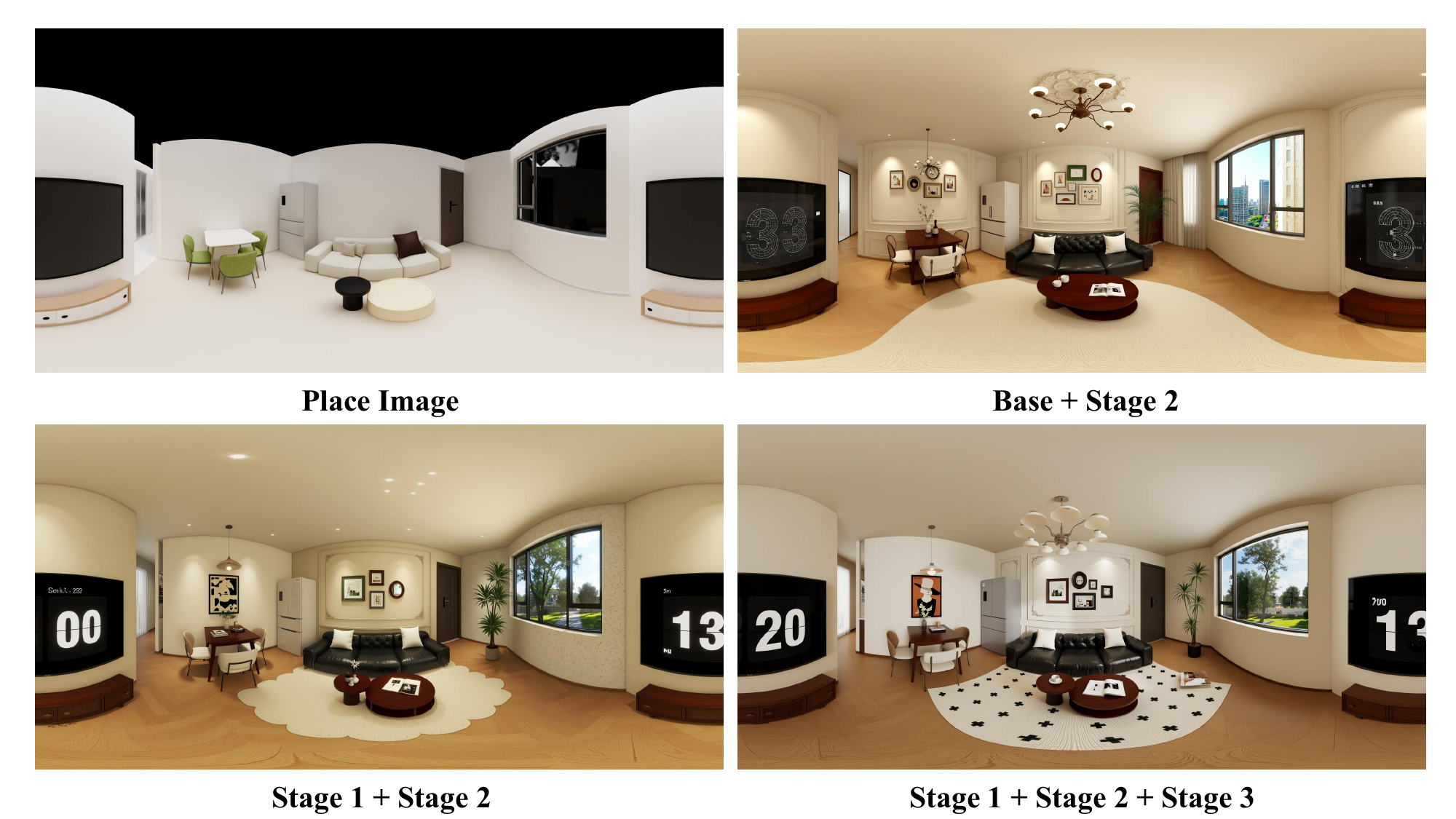} 
        \caption{living room.}
        \label{sft comparison 1}
    \end{subfigure}
    \hfill
    \begin{subfigure}[b]{0.48\textwidth}
        \centering
        \includegraphics[width=\textwidth]{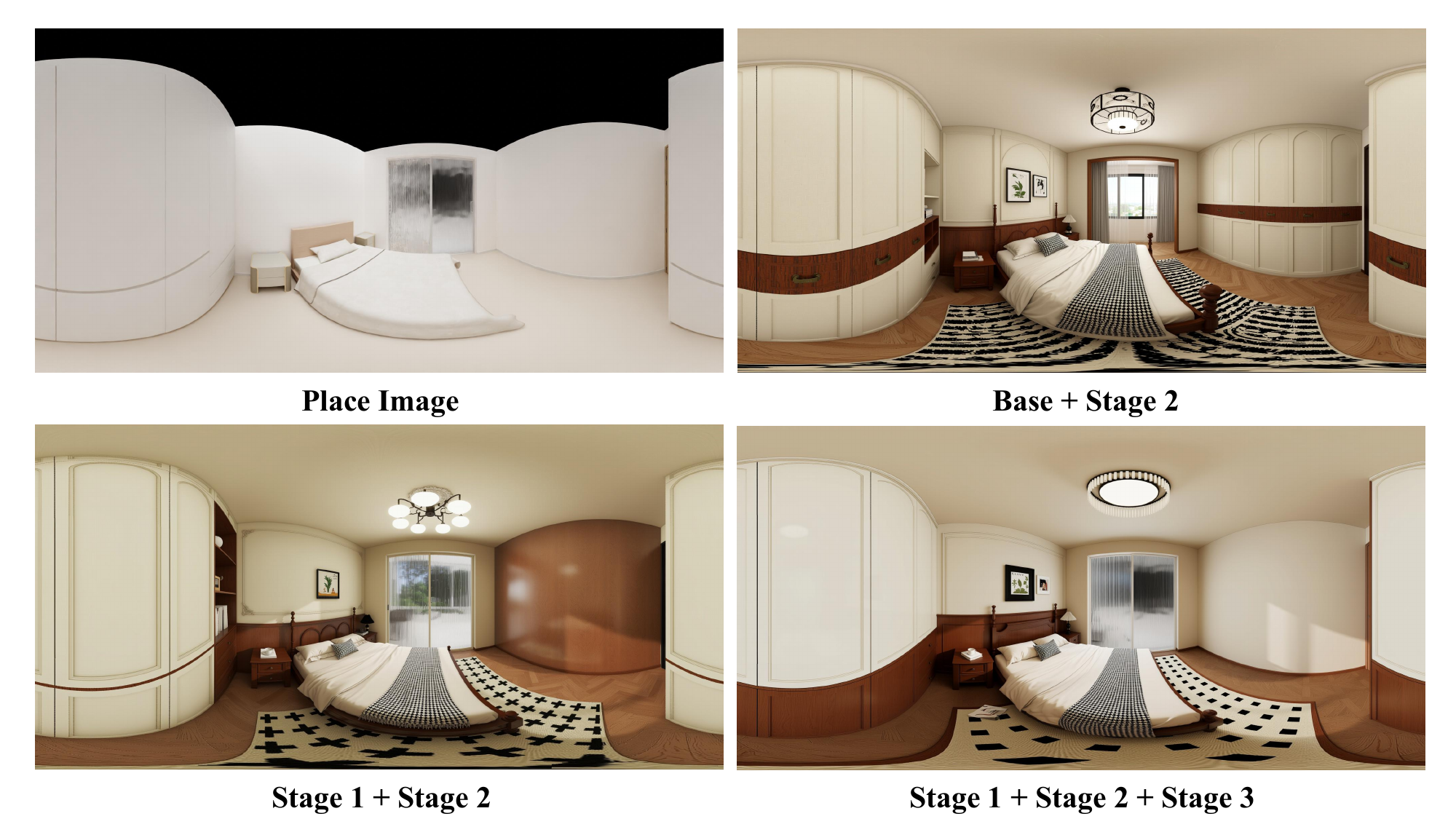} 
        \caption{bedroom.}
        \label{sft comparison 2}
    \end{subfigure}
    \caption{Qualitative ablation study of three-stage supervised fine-tuning.}
    \label{fig:sft comparison}
\end{figure}

Figures \ref{fig:sft comparison} illustrate the impact of our multi-stage training strategy on balancing spatial consistency and visual rendering quality.

\begin{itemize}
    \item \textbf{Spatial Consistency Failures (Base + Stage 2):} As shown in the top-right results, models trained without the large-scale panoramic data from Stage 1 struggle to master complex geometric priors. This leads to significant structural hallucinations, such as the erroneous generation of a redundant window in the living room or the complete disappearance of the glass sliding door in the bedroom layout.
    \item \textbf{Rendering Quality Bottlenecks (Stage 1 + Stage 2):} While incorporating Stage 1 data improves spatial layout (bottom-left), the inherent noise and uneven quality of the large-scale dataset limit the model's aesthetic performance. The resulting images often suffer from sub-optimal lighting and "muddy" textures, appearing dim and professionally unrefined.
    \item \textbf{Synergy of the Full Pipeline (Stage 1 + Stage 2 + Stage 3):} The integrated results (bottom-right) demonstrate superior performance across both dimensions. By refining the model through all three stages, we successfully correct structural anomalies while achieving high-fidelity lighting and professional-grade textures. This confirms that our multi-stage approach effectively transitions the model from basic geometric understanding to high-end aesthetic synthesis, delivering photorealistic 360-degree scenes that strictly adhere to the input architectural specifications.
\end{itemize}

\subsubsection{Reinforcement Learning}

\begin{table}[H]
    \centering
    \footnotesize  
    \caption{Quantitative ablation study of the two-stage reinforcement learning.}
    \label{tab:rl result}
    \setlength{\tabcolsep}{1.5pt}  
    \renewcommand{\arraystretch}{1.1}  
    \begin{tabular}{ccccccccccc}
        \toprule
        & \multicolumn{7}{c}{\textbf{Spatial Consistency}} & \textbf{Aesthetic}& \textbf{Similarity}& \textbf{Realism}\\
        \cmidrule(lr){2-8} \cmidrule(lr){9-9} \cmidrule(lr){10-10} \cmidrule(lr){11-11}
        & Wall & Door & Window & Cabinet & Sofa & Bed & Average & HPSv3 & CLIP score & OmniAID score\\
        \midrule
        \makecell{Base} &  0.9629&  0.6699&  0.5275&  0.5417&  0.7295&  0.7486&  0.6967&  7.1441&  0.8251&  0.1766\\
        \makecell{Base + NFT}&  0.9635&  0.6628&  0.5292&  0.5575&  0.7382&  0.7481&  0.6999&  7.3108&  0.8261&  0.1774\\
        \makecell{Base + DPO}&  0.9640&  0.6872&  0.5449&  0.5564&  0.7264&  0.7420&  0.7035&  7.3624&  0.8268&  0.1780\\
        \textbf{\makecell{Base + DPO + NFT}}  &  0.9650&  0.6770&  0.5416&  0.5594&  0.7296&  0.7489&  0.7036&  7.4648&  0.8230& 0.1793\\ 
        \bottomrule 
    \end{tabular}
\end{table}

To evaluate the impact of our post-training strategy, we analyze the learning dynamics of DPO and NFT through the lens of sample variance. As shown in Table \ref{tab:rl result}, both stages contribute to a comprehensive improvement across all metrics, including structural fidelity and aesthetic quality. However, their optimization efficiency is closely tied to the standard deviation (STD) of the training signals.

\begin{figure}[H]
    \centering
    \begin{subfigure}[b]{0.48\textwidth}
        \centering
        \includegraphics[width=\textwidth]{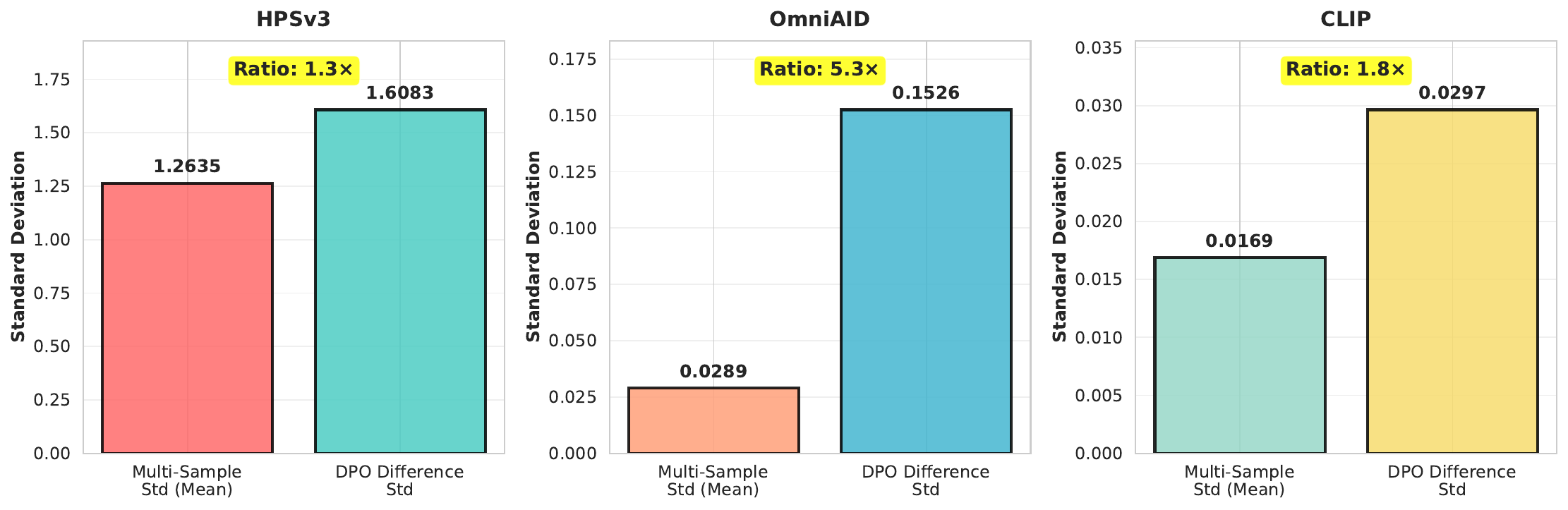} 
        \caption{Comparison of Standard Deviation between DPO and NFT sample groups.}
        \label{fig:std_comparison}
    \end{subfigure}
    \hfill
    \begin{subfigure}[b]{0.48\textwidth}
        \centering
        \includegraphics[width=\textwidth]{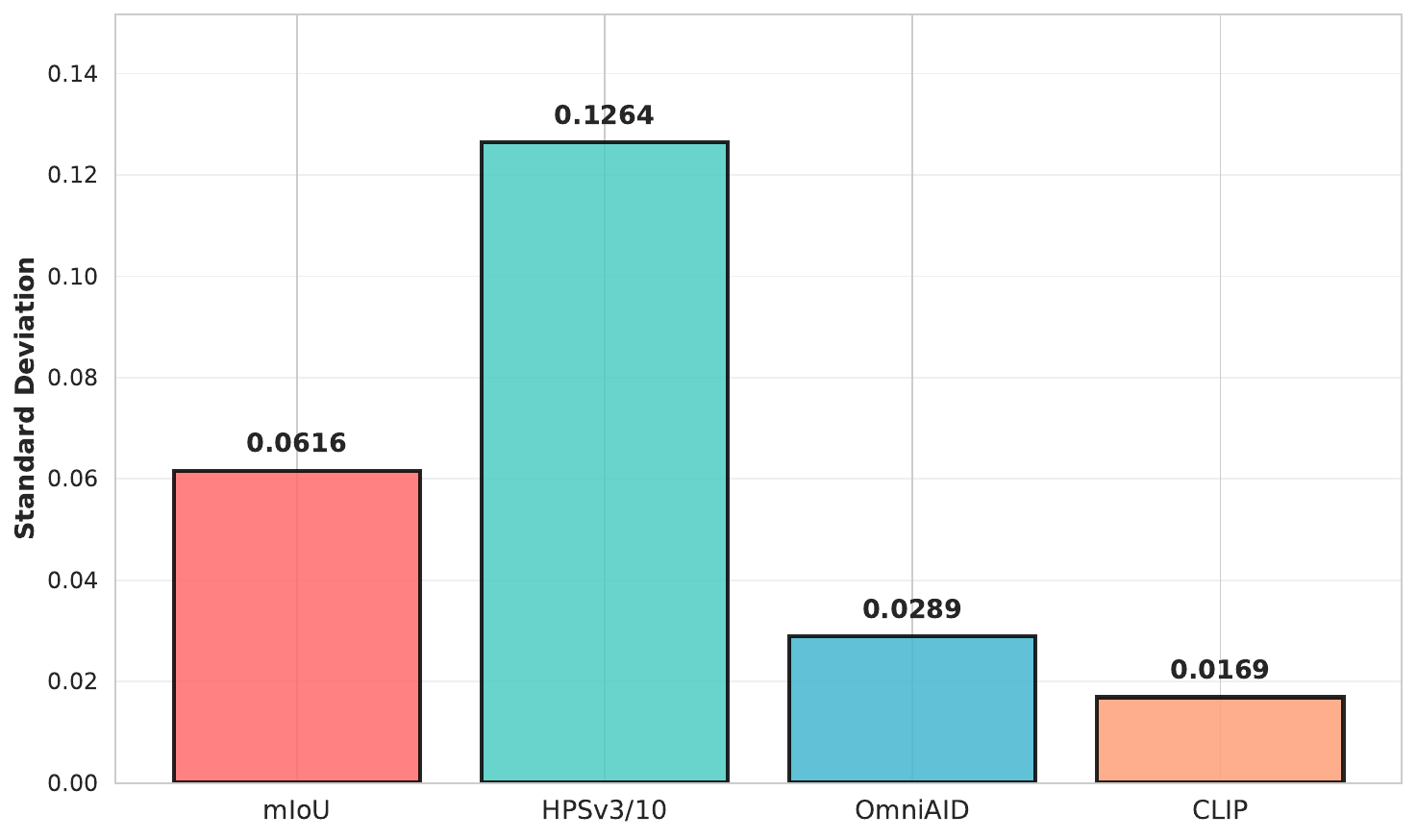} 
        \caption{Standard Deviation of the four reward components within NFT.}
        \label{fig:std_rewards}
    \end{subfigure}
    \caption{Statistical analysis of the training signals during the RL phases. (a) Highlights the diversity constraint in the multi-conditioned NFT stage; (b) Demonstrates that the HPSv3 reward provides the strongest optimization signal due to its high variance.}
    \label{fig:std_analysis}
\end{figure}

Our statistical analysis (Figure \ref{fig:std_analysis}) reveals that the standard deviation among the 1-to-N sample groups in the NFT phase is significantly lower than that of the self-contrastive pairs in the DPO phase. This is primarily because DreamHome-Pano operates under dense control conditions (layout masks, style references, and text prompts), which heavily constrain the generative policy. The resulting high similarity among sampled candidates limits the diversity of the feedback, explaining why the performance gains in the NFT stage are more smaller compared to the DPO stage.

Furthermore, we observe that among the four composite rewards used in the NFT phase, HPSv3 exhibits the highest standard deviation. This high variance indicates that aesthetic preferences provide a more discriminative signal within the constrained sample groups compared to structural or semantic metrics. Consequently, the optimization process captures a stronger gradient for visual refinement, leading to the most substantial improvements in aesthetic and realism scores. In conclusion, while the heavy conditioning of panoramic scenes naturally limits the diversity of reinforcement signals, the multi-objective reward function—particularly the high-variance aesthetic component—effectively drives the model toward professional-grade design standards.

\subsubsection{Prompt-LLM}
\begin{figure}[t]
    \centering
    \includegraphics[scale=0.4]{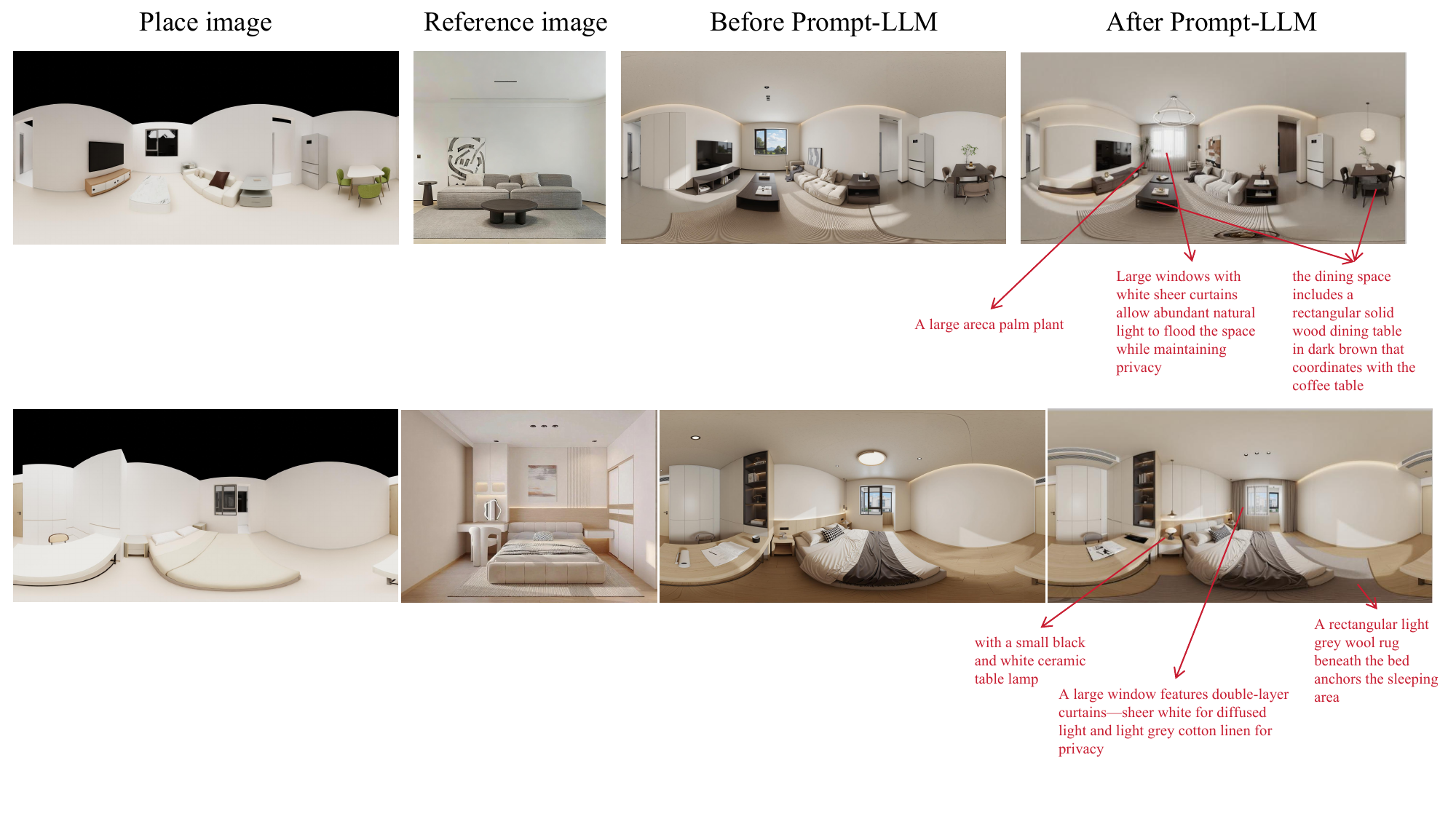}
    \caption{Qualitative comparison of semantic guidance between the rule-based baseline and Prompt-LLM.}
    \label{fig:prompt-llm-comp}
\end{figure}

To evaluate the effectiveness of the Design-Aware Prompt-LLM, we compare its performance against a rule-based baseline in Figure~\ref{fig:prompt-llm-comp} and Table~\ref{tab:prompt-llm metrics}. The baseline relies on heuristic logic to merge elements from the layout and reference images, which often results in sparse descriptions and incomplete scenes.

\begin{table}[H]
    \centering
    \footnotesize  
    \caption{Quantitative ablation study of Prompt-LLM. }
    \label{tab:prompt-llm metrics}
    \setlength{\tabcolsep}{1.0pt}  
    \renewcommand{\arraystretch}{1.1}  
    \begin{tabular}{ccccccccccc}
        \toprule
        & \multicolumn{7}{c}{\textbf{Spatial Consistency}} & \textbf{Aesthetic}& \textbf{Similarity}& \textbf{Realism}\\
        \cmidrule(lr){2-8} \cmidrule(lr){9-9} \cmidrule(lr){10-10} \cmidrule(lr){11-11}
        & Wall & Door & Window & Cabinet & Sofa & Bed & Average & HPSv3 & CLIP score & OmniAID score\\\midrule
 \makecell{Ours (w/o Prompt-LLM)}& 0.9649& 0.6757& \textbf{0.6669}& 0.5230& 0.7609& 0.7473& \textbf{0.7231}& 6.8867& 0.8277&0.1829\\ 
        
 \makecell{Ours (w/ Prompt-LLM)}& 0.9629& 0.6699& 0.5275& 0.5417& 0.7295& 0.7486& 0.6967& \textbf{7.1441}& 0.8251&0.1766\\ \bottomrule 
    \end{tabular}
\end{table}

While the rule-based approach may yield slightly higher IoU scores in some cases, this is primarily because it strictly limits the generation to a minimalist set of core furniture. In contrast, our Prompt-LLM performs \textbf{semantic expansion} by leveraging its learned design priors. For instance, it intelligently populates the scene with complementary elements—such as curtains, single-person sofas, and decorative ornaments—that are not explicitly present in the input reference but are stylistically appropriate for the target space.

While this expansion leads to a marginal decrease in IoU (as the model introduces new geometric instances not defined in the original sparse masks), it significantly enhances the \textbf{semantic richness} and \textbf{aesthetic completeness} of the panorama. As shown in the visual comparison, the Prompt-LLM successfully infers textures, materials, and lighting for core items while filling empty corners with logically inferred design elements, resulting in a more professional and holistic interior narrative compared to the rule-based baseline.

\subsubsection{Conflict-Free Control}
\begin{figure}[t]
    \centering
    \includegraphics[width=0.8\linewidth]{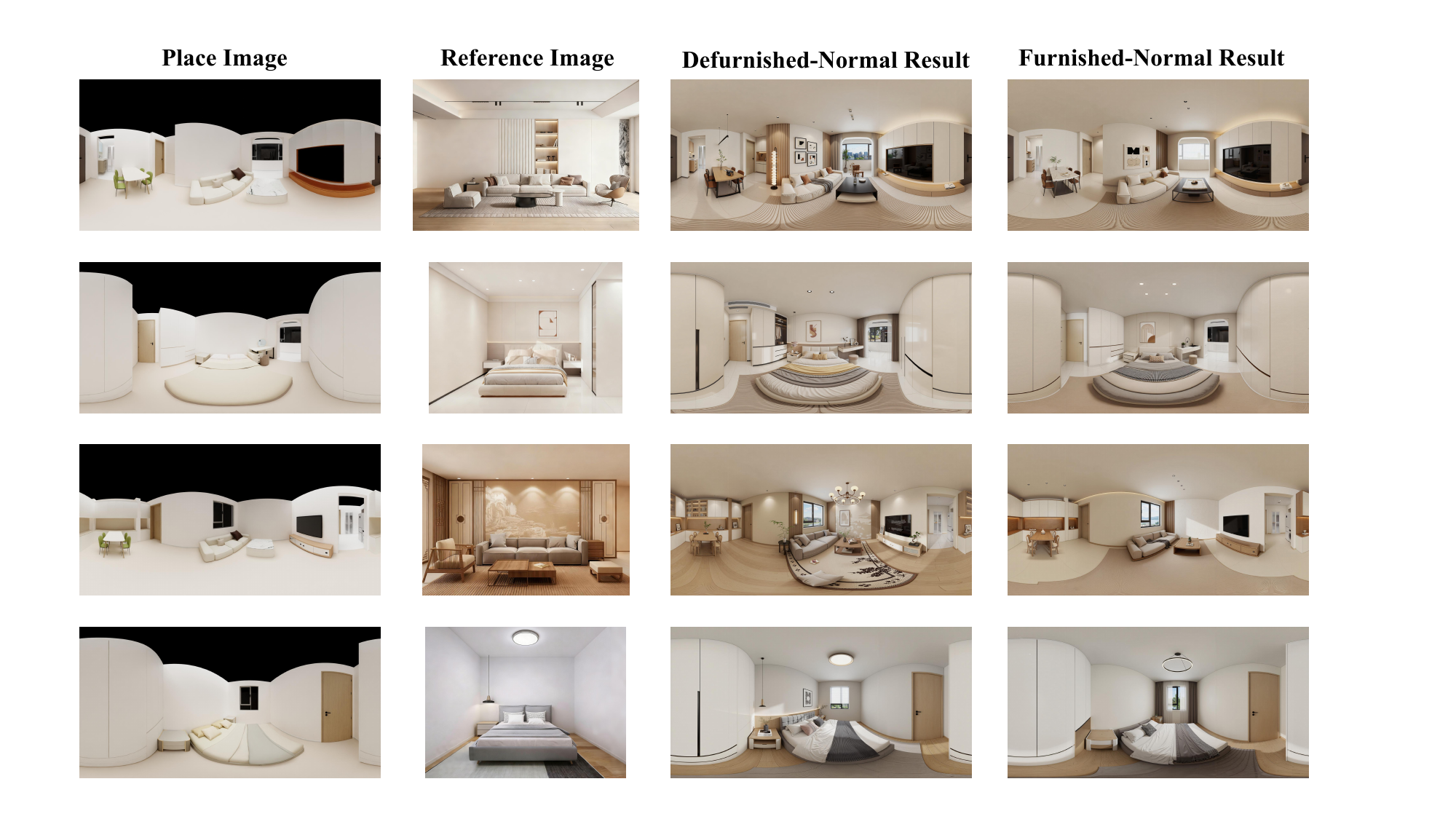}
    \caption{Visual comparison of generation results using defurnished versus furnished normal maps, demonstrating the necessity of the defurnishing module for enhancing output richness. }
    \label{fig:Qualitative vs defurnished}
\end{figure}

To evaluate the effectiveness of our defurnishing module, we compare the generation performance of models trained on normal maps derived from furnished panoramas against those trained on normal maps synthesized from empty-room panoramas, respectively. As shown in Figure \ref{fig:Qualitative vs defurnished}, using furnished normal maps produces noticeably plainer and less detailed results. This occurs because non-empty normal maps encode dense features of specific furniture and decorative elements, causing the model to overly rely on them for visual complexity. During inference, when the input place images lack such high-frequency details, the model fails to generate new decorative elements, yielding sparse and under-decorated scenes. 

In contrast, the generation results using defurnished-normal maps exhibit much better expressiveness and richness. Training on clean architectural shells allows the model to synthesize complex interior textures and furniture arrangements from design-aware prompts, rather than merely reconstructing existing geometry. These results confirm that defurnishing transforms the normal map from a rigid reconstruction template into a flexible structural prior, substantially enhancing the model’s generative capacity for professional interior design.

\begin{figure}[t]
    \centering
    \includegraphics[width=1\linewidth]{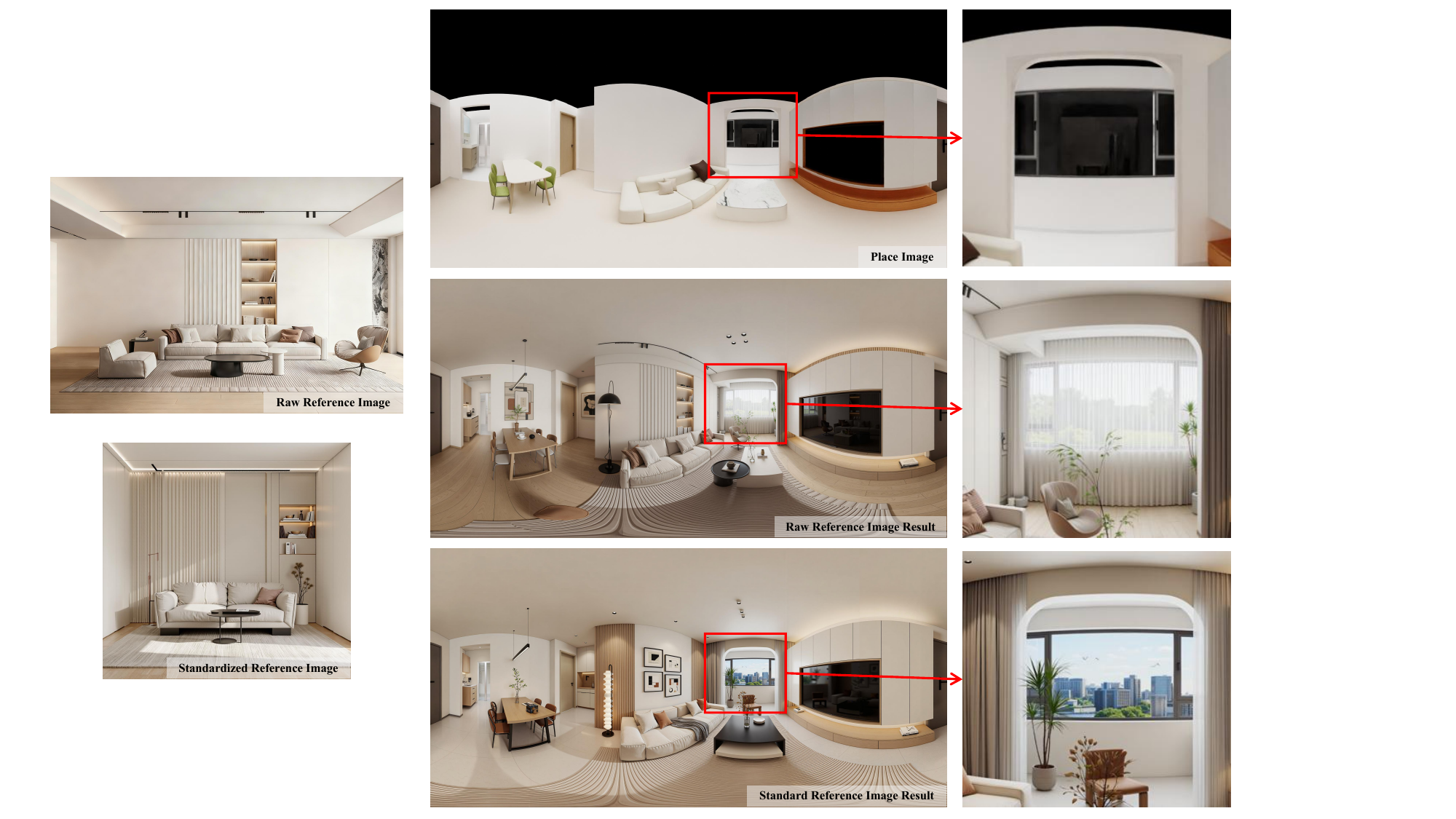}
    \caption{Qualitative Comparison of Living Room Generation: Raw Reference vs. Standardized Reference}
    \label{fig:standard ref living room}
\end{figure}

\begin{figure}[t]
    \centering
    \includegraphics[width=1\linewidth]{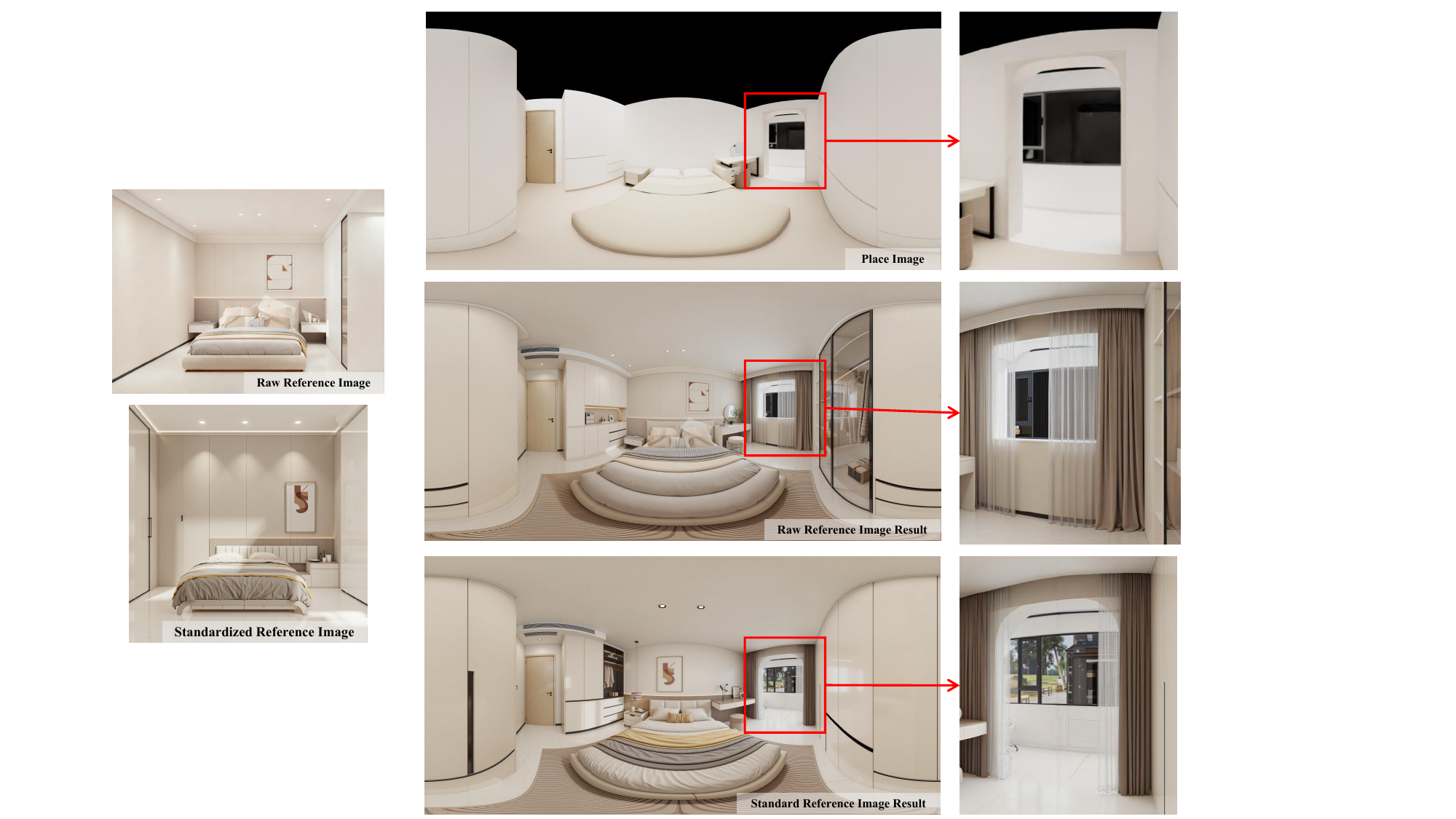}
    \caption{Qualitative Comparison of Bedroom Generation: Raw Reference vs. Standard Reference}
    \label{fig:standard ref bedroom}
\end{figure}

The qualitative comparisons in the living room and bedroom scenarios, as shown in Figure \ref{fig:standard ref living room}\ref{fig:standard ref bedroom},  illustrate the necessity of reference image standardization during the training phase in resolving the trade-off between structural fidelity and aesthetic guidance. Without standardization, raw reference images introduce native spatial noise that conflicts with the target layout, causing structural distortions and degradation of architectural elements such as window frames and wall niches.  By projecting the reference onto a neutral spatial template, our framework effectively decouples stylistic attributes from structural priors, allowing the model to adopt the color palette and lighting of the style reference while strictly preserving the geometric integrity and spatial hierarchy of the original architecture. 

\FloatBarrier

\section{Conclusion and Limitation}
In this report, we presented \textbf{DreamHome-Pano}, a novel controllable framework for high-fidelity 360-degree panoramic interior generation. Our approach successfully addresses the critical challenge of multi-condition interference by integrating a \textbf{Prompt-LLM} for professional semantic guidance and a \textbf{Conflict-Free Control Strategy} that decouples structural fidelity from stylistic transfer. Experimental results demonstrate that DreamHome-Pano achieves a superior balance between aesthetic richness and structural consistency, providing a robust technical foundation for AI-driven immersive interior design.

Despite these advancements, certain limitations remain. While our framework ensures global geometric alignment, the generative process can still produce subtle artifacts or minor distortions in complex local details. Additionally, although the Prompt-LLM significantly enhances semantic richness, occasional physical implausibility in decorative object placement may still occur. Future work will focus on further refining these micro-level details through more advanced geometric constraints and enhanced physical-aware reward functions to achieve even higher levels of photorealistic precision.

\bibliographystyle{unsrt}  
\bibliography{references}
\end{document}